%% file: main.tex
\documentclass{article}
\usepackage{iclr2027_conference,times}
\usepackage[T1]{fontenc}
\usepackage[utf8]{inputenc}
\usepackage{amsmath,amssymb,booktabs,graphicx,tabularx,array,microtype}
\usepackage{float,enumitem}
\floatstyle{ruled}
\newfloat{algorithm}{tbp}{loa}
\floatname{algorithm}{Algorithm}
\usepackage{xcolor,colortbl,hyperref,xurl,etoolbox,placeins}
\definecolor{paperblue}{HTML}{24527A}
\definecolor{gainup}{HTML}{D45D2D}
\definecolor{gaindown}{HTML}{008D9C}
\definecolor{gainflat}{HTML}{606872}
\definecolor{rowshade}{HTML}{F4F4F6}
\definecolor{headerfill}{HTML}{EDF2EA}
\definecolor{oursfill}{HTML}{E7F1E1}
\definecolor{tablegreen}{HTML}{39784D}
\newcommand{\gain}[1]{\textcolor{gainup}{\ensuremath{\uparrow\!#1}}}
\newcommand{\loss}[1]{\textcolor{gaindown}{\ensuremath{\downarrow\!#1}}}
\hypersetup{colorlinks=true,linkcolor=red,citecolor=cyan,urlcolor=magenta,
 pdftitle={Selective Transfer of RL Updates for Visual Reasoning},
 pdfauthor={Suxin Ji, Hungtao Wan, Mingjun Liu, An Zhang},pdfsubject={Experimental evaluation of selective RL-update transfer for visual reasoning}}
\newcommand{\ourmethod}{\textsc{Selective-RL}}
\title{Selective Transfer of RL Updates\\for Visual Reasoning}
\author{Suxin Ji$^{1}$\quad Hungtao Wan$^{2}$\quad Mingjun Liu$^{3}$\quad An Zhang$^{4}$ \\
$^{1}$University of Pennsylvania\quad $^{2}$Independent Researcher \\
$^{3}$University of Electronic Science and Technology of China\quad $^{4}$Chongqing University}
\iclrfinalcopy
\begin{document}
\maketitle
\lhead{Under review as a conference paper at ICLR 2027}
\begin{abstract}
Model merging provides a training-free way to transfer reasoning capabilities from language models to vision-language models (VLMs), but endpoint-based transfer can conflate pre-existing model differences with changes acquired during reasoning post-training. We instead formulate capability transfer around the training-stage update, isolating the parameter changes induced by reinforcement learning (RL). Yet transferring this update in full remains suboptimal: we find that its components differ substantially in cross-model transferability, with dominant directions transferring more effectively than the complete update. Based on this finding, we introduce Selective-RL, which isolates the RL-stage update, retains its dominant matrix-wise directions with magnitude preservation, and transfers them to the language modules of a VLM. Across three model families and five visual-reasoning benchmarks, Selective-RL improves full-update interpolation in 12 of 15 comparisons, including an 8.55 percentage-point MathVision gain on the Qwen recipient. Matched controls show that update magnitude or arbitrary low rank alone does not reproduce these gains. These results highlight a distinction between what is acquired during post-training and what remains transferable across models, providing a training-stage perspective on cross-model capability transfer. Code is available at \url{https://anonymous.4open.science/r/selective-rl}.
\end{abstract}
\input{sections/introduction}
\input{sections/related_work}
\input{sections/method}
\input{sections/experiments}

\FloatBarrier
\input{sections/discussion}
\FloatBarrier
\label{main-content-end}
\clearpage
\input{sections/statements}
\bibliography{references}
\bibliographystyle{iclr2027_conference}
\clearpage
\appendix
\input{sections/appendix}

\input{sections/case_appendix}

\end{document}

%% file: sections/introduction.tex
\section{Introduction}
\label{sec:introduction}

Model merging transfers capabilities across pretrained models without additional training~\citep{marczak2025isotropic,fang2025neuromerging}, including reasoning transfer from language models to vision-language models (VLMs)~\citep{chen2025bring,ipmerging,frism}. Endpoint-based transfer constructs a donor--recipient displacement, while parameter selection, projection, and spectral filtering address interference~\citep{gargiulo2025tsv,lee2025star,lee2026adarank}. Yet the donor endpoint alone cannot identify reasoning post-training changes. Analyses of base-model capabilities and reinforcement learning (RL) distinguish pre-existing reasoning behavior from post-training effects~\citep{liu2025drgrpo,yue2025rlcapacity,dapo}. For a reasoning donor and a multimodal recipient, their displacement can therefore mix differences that predate RL with changes acquired during RL.

This distinction becomes explicit when we track parameter provenance, as in task-vector and alignment transfer~\citep{taskarithmetic,huang2024chatvector} and interpolation between reward-specialized models~\citep{rame2023rewardedsoups}. Let $V$ denote the multimodal recipient, $B$ the donor's actual pre-RL checkpoint, and $R$ its post-trained counterpart. On their aligned language parameters,
\[
\underbrace{\mathrm{R}-\mathrm{V}}_{\text{endpoint displacement}}
=\underbrace{\mathrm{B}-\mathrm{V}}_{\text{pre-RL displacement}}
+\underbrace{\mathrm{R}-\mathrm{B}}_{\text{RL-stage update}}.
\]
Here $\mathrm{B}-\mathrm{V}$ predates RL, whereas $\mathrm{R}-\mathrm{B}$ isolates RL-acquired changes. Merging depends on parameter importance and functional mismatch~\citep{matena2022fisher,jin2023regmean,daheim2024gradientmatching}; task information can occupy localized directions or parameter supports~\citep{ortiz2023tangent,wang2024localizing}. Selective parameter transfer can also preserve existing multimodal capabilities~\citep{zhu2024modeltailor}. These findings motivate isolating the RL-stage update before testing which components transfer; stage attribution alone does not establish that every component encodes transferable reasoning.

RL-stage isolation is necessary but insufficient: $\mathrm{R}-\mathrm{B}$ was optimized for the source language model, so its components may differ in usefulness to a multimodal recipient. This question is particularly relevant because RL updates in reasoning models exhibit pronounced spectral structure, with dominant directions capturing a disproportionate part of their learning dynamics~\citep{cai2026predictability}. Prior work primarily examines this structure through source-model reconstruction or trajectory characterization; whether it identifies components compatible across models remains unclear. In the primary Qwen controls, the dominant matrix-wise direction outperforms the complete update at matched Frobenius magnitude; retaining more directions progressively reduces MathVista accuracy. Thus, components of the same post-training update can differ substantially in cross-model transferability, and preserving more of the source update need not be beneficial.

Motivated by this finding, we introduce \textbf{Selective-RL}, a training-free framework for transferring reasoning capabilities from RL-trained language models to multimodal recipients. It first extracts the RL-stage update $\mathrm{R}-\mathrm{B}$, excluding parameter differences that predate reasoning post-training. For each eligible language-layer matrix, it then retains the dominant singular direction and rescales it to preserve the magnitude of the original matrix-wise update. We compare direct injection and reconstructed-donor interpolation on aligned language modules, preserving visual components. The interpolation comparison holds pre-RL displacement fixed. Unlike multimodal post-training approaches~\citep{deng2025openvlthinker,huang2026visionr1,wang2025vlrethinker}, Selective-RL requires no multimodal post-training, recipient gradients, additional modules, or target-side adaptation, and preserves the VLM's architecture and inference structure.

We evaluate Selective-RL on three multimodal recipients spanning the Qwen~\citep{qwen3vl}, LLaMA~\citep{grattafiori2024llama3}, and Mistral~\citep{jiang2023mistral} families across five visual-reasoning benchmarks. It improves matched full-update interpolation in \textbf{12/15} backbone--benchmark comparisons (Table~\ref{tab:T1}). On Qwen3-VL, MathVision rises \textbf{$40.46\to49.01\%$} (\textbf{$+8.55$ pp}), MathVerse $45.30\to52.00\%$ ($+6.70$ pp), and MathVista $76.70\to81.00\%$ ($+4.30$ pp). On MathVista, norm-matched leading directions reach \textbf{$81.00\%$}, versus $74.05\pm0.45\%$ for ten matched random-direction runs and $73.10\%$ for trailing directions (Table~\ref{tab:T3}). Increasing rank from one to sixteen lowers accuracy to $78.30\%$, while gains over Full average $4.13\pm0.15$ pp across three donor seeds (Tables~\ref{tab:T4} and~\ref{tab:donor_seed}). These results distinguish source-update preservation from recipient improvement: selecting transferable structure can be more effective than retaining the complete update, with task-dependent exceptions.

%% file: sections/related_work.tex
\section{Related Work}
\label{sec:related}

\noindent\textbf{Model Merging.}
Task Arithmetic~\citep{taskarithmetic} composes checkpoint differences. TIES-Merging~\citep{ties} trims updates and resolves sign disagreement; DARE~\citep{dare} drops and rescales delta parameters. TSV~\citep{gargiulo2025tsv} compresses task matrices and decorrelates singular vectors. LoRE-Merging~\citep{liu2025lore} estimates a reference and low-rank task vectors without the original base. STAR~\citep{lee2025star} truncates by cumulative spectral mass and restores nuclear norms. AdaRank~\citep{lee2026adarank} learns component masks by entropy minimization on unlabeled test inputs. We compare adaptive selection and scaling with fixed leading-component transfer of the actual RL update under Frobenius matching.

\medskip
\noindent\textbf{Multimodal Reasoning.}
OpenVLThinker~\citep{deng2025openvlthinker} alternates supervised fine-tuning and RL. Vision-R1~\citep{huang2026visionr1} combines a reasoning cold start, GRPO, and Progressive Thinking Suppression Training. VL-Rethinker~\citep{wang2025vlrethinker} uses Selective Sample Replay and Forced Rethinking. Bring Reason to Vision~\citep{chen2025bring} merges language parameters while retaining visual components. IP-Merging~\citep{ipmerging} identifies, rescales, and projects reasoning-associated parameters into recipient subspaces. FRISM~\citep{frism} learns subspace coefficients through label-free multimodal self-distillation. We separate RL updates from initialization displacement without recipient gradients, retaining validation-selected coefficients.

\medskip
\noindent\textbf{RL Updates.}
Source recovery, generation behavior, and recipient improvement require separate evidence. DeepSeek-R1~\citep{deepseek2025r1} contrasts direct RL with multistage training, making the reference checkpoint essential. DAPO~\citep{dapo} studies decoupled clipping and dynamic sampling. Dr.\ GRPO~\citep{liu2025drgrpo} examines base-model behavior and response-length biases. Large-budget pass@$k$ analysis~\citep{yue2025rlcapacity} distinguishes sampling efficiency from solution coverage. \citet{cai2026predictability} study low-rank RL reconstruction and introduce AlphaRL for trajectory extrapolation. Frankenstein-style analysis~\citep{li2026frankenstein} probes visual-RL refinements through causal interventions, checkpoint exchange, and freezing. We control direction, magnitude, and initialization displacement; spectral concentration alone identifies neither causal mechanisms nor RL-exclusive semantic components.

%% file: sections/method.tex
\section{Methodology}
\label{sec:method}

Our goal is to transfer the capability acquired during the donor's RL stage while separating it from parameter differences that predate RL. We isolate the RL-stage update from initialization displacement (Section~\ref{sec:3.1}), select its dominant matrix-wise direction at controlled magnitude (Section~\ref{sec:3.2}), and compose it with the recipient through two transfer routes (Section~\ref{sec:3.3}). Section~\ref{sec:3.4} characterizes update scale and resource use.

\subsection{Problem Definition and Stage Isolation} \label{sec:3.1}

We begin by separating two sources of parameter change that are mixed in conventional donor--recipient merging. Let \(V=(V_j)_{j\in\mathcal I}\) denote the multimodal recipient, \(B\) the actual checkpoint from which RL training starts, and \(R\) the resulting reasoning donor. Let \(\mathcal S\subseteq\mathcal I\) contain the transferable attention and MLP matrices that are aligned between donor and recipient, with \(V_j,B_j,R_j\in\mathbb{R}^{m_j\times n_j}\) for \(j\in\mathcal S\). We define \(\Gamma_j:=B_j-V_j\) and \(\Delta_j:=R_j-B_j\) for \(j\in\mathcal S\).

Here, \(\Gamma_j\) captures parameter differences that already exist before RL, whereas \(\Delta_j\) isolates the update acquired during RL. This separates RL-acquired changes from pre-existing donor--recipient mismatch.

The support \(\mathcal S\) contains the aligned \(q/k/v/o\) projections in attention blocks and the gate/up/down projections in MLP blocks, matched by layer and module with consistent orientation and head ordering. Parameters outside \(\mathcal S\), including the recipient's visual components, remain unchanged.

To express different transfer strategies in a common form, we define zero extension for any collection \(H=(H_j)_{j\in\mathcal S}\), \([\mathcal E_{\mathcal S}(H)]_j=H_j\) for \(j\in\mathcal S\) and \(0\) otherwise, and write the merged model as
\begin{equation}
 W(\alpha,\beta;Z) = V+\mathcal E_{\mathcal S}(\alpha\Gamma+\beta Z),
 \label{eq:unified}
\end{equation}
where \(Z\) is the RL-derived signal to be transferred. The coefficients \(\alpha\) and \(\beta\) independently control the initialization displacement and the RL-stage contribution, enabling us to separate their effects.

Conventional interpolation~\citep{chen2025bring} is recovered by setting \(Z=\Delta\) and \(\alpha=\beta=\lambda\):
\begin{equation}
 W^{\mathrm{full}}_j = V_j+\lambda\Gamma_j+\lambda\Delta_j = (1-\lambda)V_j+\lambda R_j, \qquad j\in\mathcal S .
 \label{eq:full-interpolation}
\end{equation}
Equation~\eqref{eq:full-interpolation} makes the coupling explicit: standard interpolation simultaneously moves the recipient toward the donor's pre-RL initialization through \(\Gamma\) and transfers the entire RL-stage update through \(\Delta\). Isolating \(\Delta\) identifies what was acquired during RL; the remaining question is whether all of this update should be transferred.

\begin{figure}[!htbp]
\centering
\includegraphics[width=\linewidth]{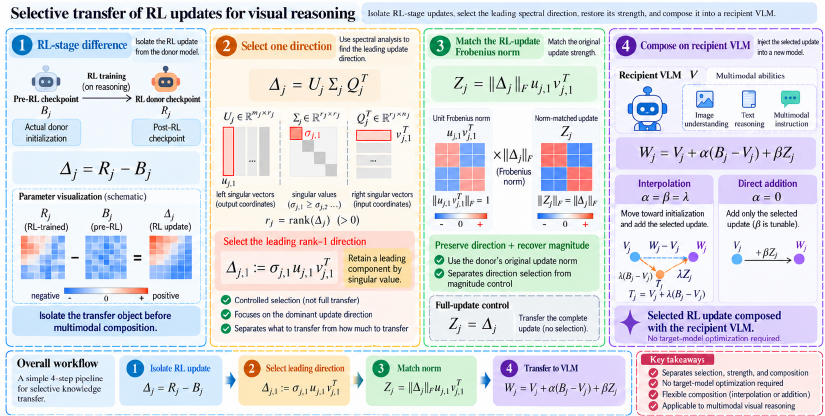}
\caption{\textbf{Selective RL-update transfer.} Four stages isolate $R-B$, select a matrixwise direction, restore its Frobenius norm, and compose it with the recipient. Interpolation includes initialization displacement; direct addition omits it. Matrix patterns are schematic.}
\label{fig:overview}
\end{figure}

Figure~\ref{fig:overview} summarizes the stage-isolation, direction-selection, normalization, and composition steps.

\subsection{Fixed Matrixwise Spectral Selection} \label{sec:3.2}

Having isolated the RL-stage update, we next ask \textbf{which part of it transfers effectively}. The complete update \(\Delta_j\) was optimized for the source language model, so its constituent directions need not be equally compatible with a multimodal recipient. We therefore decompose each update matrix independently and study its spectral directions under controlled magnitude.

For a nonzero \(\Delta_j\) with rank \(r_j\), the compact singular value decomposition is \(\Delta_j=U_j\Sigma_jQ_j^\top\), where \(U_j=[u_{j,1},\ldots,u_{j,r_j}]\) and \(Q_j=[v_{j,1},\ldots,v_{j,r_j}]\) contain the left and right singular vectors, respectively, and \(\sigma_{j,1}\geq\cdots\geq\sigma_{j,r_j}>0\), with \(U_j^\top U_j=Q_j^\top Q_j=I_{r_j}\). For a positive integer \(k\), we retain the leading \(k\) singular components, \(\mathcal P_k(\Delta_j):=\Delta_{j,k}=\sum_{i=1}^{\min(k,r_j)}\sigma_{j,i}u_{j,i}v_{j,i}^{\top}\), with $\mathcal P_k(0)=0$ and empty sums equal to zero. For nonzero updates, these approximations satisfy
\begin{equation}
 \|\Delta_j-\Delta_{j,k}\|_F^2 = \min_{\operatorname{rank}(A)\leq k} \|\Delta_j-A\|_F^2.
 \label{eq:truncation-residual}
\end{equation}
Equation~\eqref{eq:truncation-residual} characterizes optimal low-rank reconstruction of the source update, which need not maximize transfer to another model. Spectral ordering defines controlled candidate directions whose transferability we evaluate empirically.

A second issue is that truncation changes both direction and magnitude. To avoid confounding these effects, we measure the fraction of squared Frobenius energy retained by the rank-\(k\) approximation, \(\rho_{j,k}:=\|\Delta_{j,k}\|_F^2/\|\Delta_j\|_F^2\) for \(\Delta_j\ne0\), and rescale the truncated update to match the Frobenius norm of the original matrix~\citep{cai2026predictability}: \(\widehat{\Delta}_{j,k}:=\mathcal N_k(\Delta_j)\), \(\mathcal N_k(\Delta_j)=\rho_{j,k}^{-1/2}\Delta_{j,k}\) for \(\Delta_j\ne0\), and \(\mathcal N_k(0)=0\). Thus, \(\|\widehat{\Delta}_{j,k}\|_F=\|\Delta_j\|_F\). The full and selected updates therefore have the same matrix-wise Frobenius magnitude while differing in directional content. Unless otherwise specified, Selective-RL uses \(k=1\) independently for each matrix, giving, for \(\Delta_j\ne0\), \(\widehat{\Delta}_{j,1}=\|\Delta_j\|_F u_{j,1}v_{j,1}^{\top}\).

Rank one is a fixed empirical rule, not an assumption that one direction constitutes reasoning capability. We vary rank, direction, and normalization separately to assess their effects on transfer.

\subsection{Controlled Transfer and Composition} \label{sec:3.3}

Spectral selection determines what within the RL update is transferred. We next separate this choice from \textbf{how the update is composed with the recipient}, since conventional interpolation also introduces the initialization displacement \(\Gamma\).

We first consider reconstructed-donor interpolation, which keeps the initialization displacement identical between full and selected transfer. Setting \(Z=\widehat{\Delta}_1\) and \(\alpha=\beta=\lambda\) gives \mbox{\(W^{\mathrm{rank1}}_j=V_j+\lambda(\Gamma_j+\widehat{\Delta}_{j,1})\)}, and \(W^{\mathrm{rank1}}_j-W^{\mathrm{full}}_j=\lambda(\widehat{\Delta}_{j,1}-\Delta_j)\).

Both variants share initialization displacement \(\lambda\Gamma_j\) and differ only in the RL contribution (\(\Delta_j\) or \(\widehat{\Delta}_{j,1}\)), providing a controlled comparison of spectral selection.

We additionally consider direct RL-update transfer by setting \(\alpha=0\): \(W^{\mathrm{direct}}_j=V_j+\beta\widehat{\Delta}_{j,1}\), \(j\in\mathcal S\).

This isolates RL-update transfer to test whether selection remains beneficial without initialization displacement.

For both routes, we compare \(Z=\widehat{\Delta}_1\) against the corresponding full-update control \(Z=\Delta\), holding transfer support, numerical precision, and coefficients fixed. Under direct transfer, \(Z=\Delta\) reduces to RL-update Task Arithmetic~\citep{taskarithmetic}. An initialization-only control instead sets \(\beta=0\) while retaining \(\alpha\Gamma\). Together, these routes separate three otherwise entangled factors: \textbf{stage isolation, spectral selection, and update composition}. Construction requires no recipient gradient training; the coefficient is selected on H1 held-out validation and frozen for H2/H3. Algorithm~\ref{alg:selective-transfer} in Appendix~\ref{sec:configuration} materializes the update.

\subsection{Update Scale and Resource Use} \label{sec:3.4}

Frobenius matching controls the magnitude of the \textbf{RL component}, but not necessarily the total displacement from the recipient. Under interpolation, the selected update also interacts with \(\Gamma\). Let \(\langle A,H\rangle_F:=\operatorname{tr}(A^\top H)\). From Eq.~\eqref{eq:unified},
\begin{equation}
 \|W_j(\alpha,\beta;Z)-V_j\|_F^2=\alpha^2\|\Gamma_j\|_F^2+\beta^2\|Z_j\|_F^2+2\alpha\beta\langle\Gamma_j,Z_j\rangle_F.
 \label{eq:displacement-norm}
\end{equation}
Matching \(\|Z_j\|_F\) therefore controls matrix-wise intervention strength while preserving directional differences through the cross term.

For rank-one selection with \(\Delta_j\ne0\), \(\|\widehat{\Delta}_{j,1}\|_2/\|\Delta_j\|_2=\rho_{j,1}^{-1/2}\geq1\). Thus, Frobenius matching amplifies the retained direction by an amount determined by the spectral concentration of the original update. This motivates separate controls for rank, direction, global coefficient, and matrix-wise rescaling rather than attributing performance changes to selection alone.

All decomposition, normalization, coefficient selection, and checkpoint construction are performed offline. The checkpoint remains a standard dense VLM without extra inference-time modules.

%% file: sections/experiments.tex
\section{Experiments}
\label{sec:experiments}
We evaluate \textbf{RQ1}, method and backbone transfer; \textbf{RQ2}, direction, scale, and composition; \textbf{RQ3}, source reconstruction and transfer boundaries; and \textbf{RQ4}, output behavior and construction cost.

\subsection{Experimental Setup}
\paragraph{Models and selection.}
H1 uses \textbf{Qwen3-VL-8B-Instruct}, H2 \textbf{LLaVA-NeXT-LLaMA3-8B}, and H3 \textbf{Idefics2-8B}, with Qwen3, LLaMA3, and Mistral language backbones~\citep{qwen3vl,llavanextmodel,idefics2}. Matched DAPO-style donors are trained on DAPO-Math-17K~\citep{dapo}, starting from the actual pre-RL initialization. With primary seed 42, a 150-step budget, and checkpoints every ten steps, source-side validation selects steps \textbf{120/100/110} for H1/H2/H3; VLM benchmarks are not used for donor selection. H1 held-out validation selects $\lambda=0.05$, which is frozen for H2/H3. The support contains $252/224/224$ attention/MLP matrices, respectively, and excludes visual components, embeddings, output heads, LayerNorm, and biases. All groups use bf16 and primary rank 1 (Appendix~\ref{sec:configuration}).

\paragraph{Tasks and comparisons.}
The primary tasks are MathVista ($N=1{,}000$), MathVerse ($1{,}000$), MathVision ($304$), DynaMath ($5{,}010$), and MMStar ($1{,}500$)~\citep{mathvista,mathverse,mathvision,dynamath,mmstar}. Additional H1 evaluations cover TextVQA, MMMU, OCRBench, and POPE~\citep{textvqa,mmmu,ocrbench,pope}. We compare \textbf{Task Arithmetic}, \textbf{Full interpolation}, \textbf{TIES}, \textbf{DARE}, \textbf{TSV}, \textbf{LoRE}, \textbf{IP-Merging}, \textbf{FRISM}, \textbf{STAR}, and \textbf{AdaRank}~\citep{taskarithmetic,chen2025bring,ties,dare,gargiulo2025tsv,liu2025lore,ipmerging,frism,lee2025star,lee2026adarank}. Full shares the proposed method's support. FRISM and AdaRank additionally adapt on the target side. We report single-run scores, donor seeds, and random-direction trials separately.

\subsection{Comparing Methods and Backbones (RQ1)}
\input{tables/T1}

\begin{figure}[!htbp]
\centering
\includegraphics[width=\linewidth]{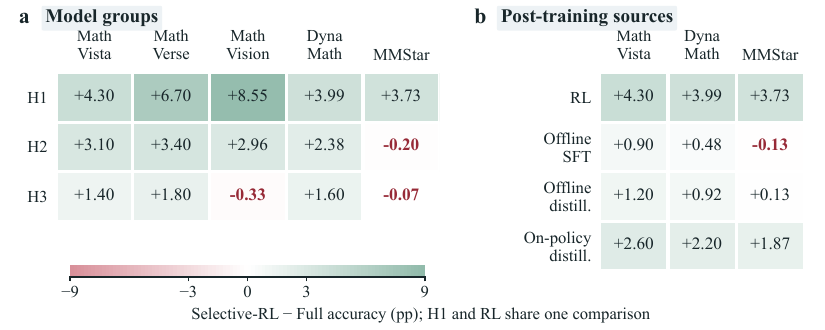}
\caption{\textbf{Transfer scope and its limits.} Rank-1 minus Full in percentage points: three receiver backbones (left) and four post-training sources on H1 (right). Negative cells retain regressions; the H1/RL setting is shared across panels.}
\label{fig:transfer-scope}
\end{figure}

\noindent\textbf{Observation 1: Gains transfer across backbones, with task-dependent exceptions.}
\ourmethod{} improves Full in \textbf{12/15} group--task comparisons. H1 gains are \gain{4.30}/\gain{6.70}/\gain{8.55}/\gain{3.99}/\gain{3.73} points from paired counts. On H1 MathVista, $81.00>79.70>79.10>76.70$ for Selective, AdaRank, STAR, and Full. On MMStar, AdaRank leads ($60.07>59.67$). H2 improves four tasks but loses $0.20$ MMStar points; H3 improves MathVista/MathVerse/DynaMath by $1.40/1.80/1.60$, while MathVision/MMStar decline $0.33/0.07$. These reversals are included in Table~\ref{tab:T1} and Figure~\ref{fig:transfer-scope}.

\noindent\textbf{Observation 2: Broader capability gains are substantial but not uniform.}
H1 gains over Full are $2.58/3.60/0.70/0.50$ points on TextVQA/MMMU/OCRBench/POPE. Relative to Receiver, OCRBench changes $78.60\to78.70$ and POPE $86.90\to86.80$. The broader scores appear in Table~\ref{tab:T10}.

\subsection{Isolating Selection and Composition (RQ2)}
\begin{figure}[!htbp]
\centering
\includegraphics[width=\linewidth]{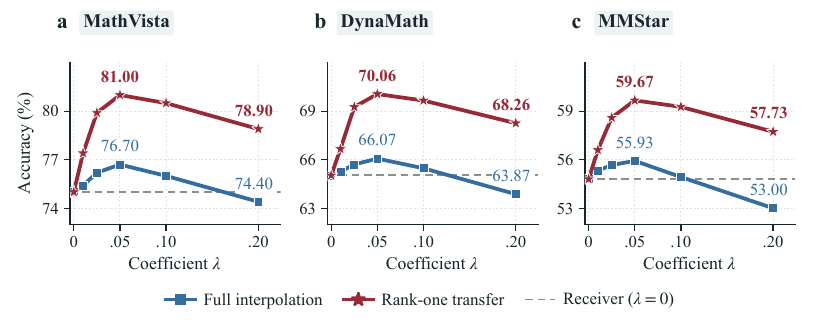}
\vspace{-1.5em}
\caption{\textbf{Coefficient sensitivity on H1.} Full and rank-one interpolation for $0\leq\lambda\leq0.2$; dashed lines mark Receiver accuracy. Validation selects $0.05$; full sweeps: Table~\ref{tab:T5}.}
\label{fig:coefficient-tradeoffs}
\end{figure}
\noindent\textbf{Observation 3: Selection helps under both composition routes.}
At $\beta=0.05$, direct/interpolated selection gains $3.10/4.30$ MathVista points and $5.70/6.70$ MathVerse points over matched Full (Table~\ref{tab:T3}). Initialization displacement lowers MathVista to $74.70$ alone but raises selected direct transfer $79.20\to81.00$, depending on the paired update. Increasing Full's coefficient does not close the gap: at $\lambda=0.2$, Rank-1 $78.90>$ Receiver $75.00>$ Full $74.40$ (Figure~\ref{fig:coefficient-tradeoffs}).

\input{tables/T3}
\begin{figure}[!htbp]
\centering
\includegraphics[width=\linewidth]{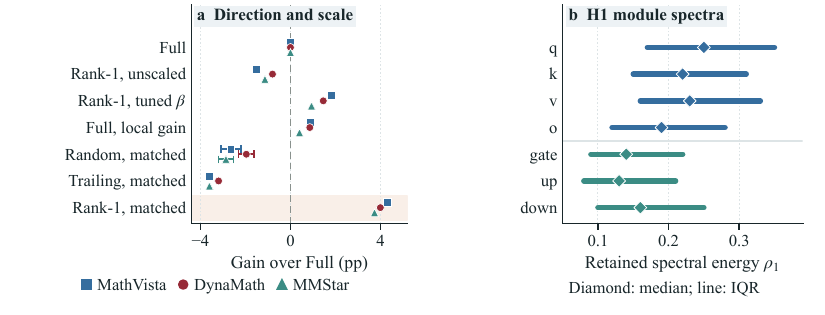}
\caption{\textbf{Direction, scale, and spectral concentration.} Left: gains over Full under the H1 controls; only random-direction means carry sample-SD bars from ten runs. Right: module-wise median retained energy $\rho_1$ and interquartile ranges across eligible matrices.}
\label{fig:direction-controls}
\end{figure}
\noindent\textbf{Observation 4: Direction selection adds value beyond a tuned global or matrixwise scale.}
Unscaled rank-1 with validation-tuned $\beta=0.50$ reaches MathVista $78.50$; applying rank-one-derived matrixwise gains to Full reaches $77.60$, versus matched selection $81.00$. Ten matched random directions yield $74.05\pm0.45$, below Receiver $75.00$; the trailing direction reaches $73.10$. On DynaMath, tuned unscaled/full-reweighted/matched-leading give $67.52/66.93/70.06$. These controls separate direction from global and matrixwise scale; semantic attribution requires separate analysis.

\begin{figure}[H]
\centering
\setlength{\abovecaptionskip}{4pt}
\setlength{\belowcaptionskip}{0pt}

\includegraphics[
    width=0.85\linewidth,
    trim={0bp 6bp 0bp 3bp},
    clip
]{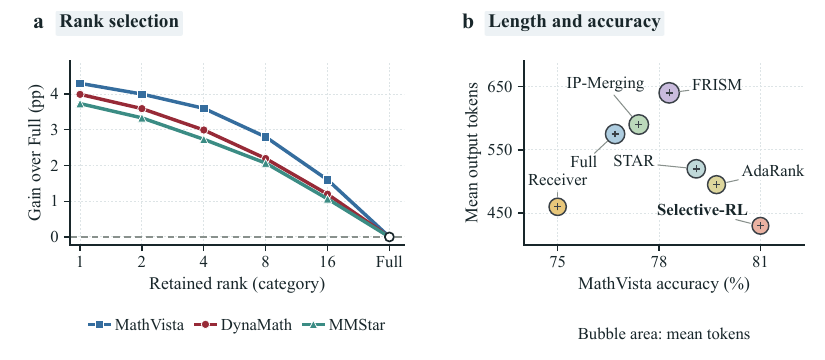}

\caption{\textbf{Rank and output efficiency.}
Left: gains over Full for finite ranks on three H1 tasks.
Right: MathVista accuracy versus mean output tokens;
bubble area encodes mean tokens.
Points show measurements without fitted trends.}
\label{fig:selection-efficiency}
\end{figure}

H1 median $\rho_1$ is $0.18$ (IQR $0.11$--$0.31$), with median amplification $2.36\times$; H2/H3 medians are $0.13/0.09$. Attention retains more leading-component energy than MLP ($0.24>0.15$ in H1). Rank 1 gives MathVista $81.00$, followed by $80.70/80.30/79.50/78.30$ at ranks $2/4/8/16$ (Figure~\ref{fig:selection-efficiency}). The rank-1/rank-2 gap is $0.30$; fixed rank one is an empirical choice.

\subsection{Source Reconstruction and Transfer Boundaries (RQ3)}

\noindent\textbf{Observation 5: Source recovery and recipient gains are distinct measurements.}
H1 rank-one reconstruction attains MATH $75.80$ versus donor $76.20$, recovering $97.18\%$ of its gain over $62.00$ pre-RL (Table~\ref{tab:T2}). The earlier-reference control lowers selective MathVista $81.00\to77.80$, supporting explicit stage isolation. Three donor seeds give MathVista gains $4.10/4.30/4.00$ ($4.13\pm0.15$). Offline SFT/distillation/on-policy distillation gain $0.90/1.20/2.60$ MathVista points; SFT loses $0.13$ MMStar points. The effect is not exclusive to RL.

Primary DynaMath gains 200 correct variants ($3.99$ points), with a reported seed-cluster 95\% interval $[2.10,5.86]$; all-ten-correct rises $29.1\to34.3\%$ (Appendix~\ref{sec:paired-outcomes}).

\subsection{Generation Behavior and Cost (RQ4)}

\noindent\textbf{Observation 6: Shorter outputs accompany gains beyond obvious generation failures.}
Mean output length falls $575\to430$ tokens (\loss{25.22\%}); repetition/extraction-failure/cap-hit counts fall $15/8/14\to3/2/3$. Of 83 Selective-only MathVista successes, $8/6/4$ correspond to Full repetition/truncation/extraction pathology and $65$ ($78.3\%$) to ordinary reasoning or perception errors. Mechanism attribution requires separate analysis. In the 200-example evaluator audit per model, disagreement is $3.0/3.5/2.5\%$ for Receiver/Full/Selective.

Construction costs $90=55+35$ s for decomposition and writing, versus Full's $35$ s, STAR's $395$ s, and AdaRank's $2{,}045$ s. Validation search is separate ($3.0$ GPU-h for Full and Selective). The dense parameter ratio remains $1.00\times$; no inference module is added (Appendix~\ref{sec:resources}).

\FloatBarrier
\input{sections/case_study}

%% file: tables/T1.tex
\begin{table}[t]
\centering
\caption{\textbf{Cross-backbone transfer and paired H1 outcomes.} Scores (\%) for methods evaluated on all five benchmarks; extended comparisons: Table~\ref{tab:extended_main}. Bold/underline: best/second within displayed groups; arrows: changes from Receiver.}
\label{tab:T1}
\begingroup
\fontsize{8.5}{10.5}\selectfont
\setlength{\tabcolsep}{9.59pt}
\renewcommand{\arraystretch}{1.10}
\begin{tabular}{lrrrrr}
\toprule
\rowcolor{headerfill}
Method & \shortstack{Math\\Vista} & \shortstack{Math\\Verse} & \shortstack{Math\\Vision} & DynaMath & MMStar \\
\midrule
\rowcolor{headerfill}\multicolumn{6}{c}{\textbf{H1: Qwen3-VL-8B-Instruct}} \\
Receiver & $75.00$ & $42.00$ & $37.50$ & $65.03$ & $54.80$ \\
\rowcolor{rowshade}
Task Arithmetic & $76.10_{\textcolor{gainup}{\uparrow1.10}}$ & $44.50_{\textcolor{gainup}{\uparrow2.50}}$ & $39.47_{\textcolor{gainup}{\uparrow1.97}}$ & $65.67_{\textcolor{gainup}{\uparrow0.64}}$ & $54.67_{\textcolor{gaindown}{\downarrow0.13}}$ \\
Full interpolation & $76.70_{\textcolor{gainup}{\uparrow1.70}}$ & $45.30_{\textcolor{gainup}{\uparrow3.30}}$ & $40.46_{\textcolor{gainup}{\uparrow2.96}}$ & $66.07_{\textcolor{gainup}{\uparrow1.04}}$ & $55.93_{\textcolor{gainup}{\uparrow1.13}}$ \\
\rowcolor{rowshade}
TIES & $76.40_{\textcolor{gainup}{\uparrow1.40}}$ & $44.80_{\textcolor{gainup}{\uparrow2.80}}$ & $39.80_{\textcolor{gainup}{\uparrow2.30}}$ & $65.77_{\textcolor{gainup}{\uparrow0.74}}$ & $55.53_{\textcolor{gainup}{\uparrow0.73}}$ \\
DARE & $76.30_{\textcolor{gainup}{\uparrow1.30}}$ & $45.00_{\textcolor{gainup}{\uparrow3.00}}$ & $40.13_{\textcolor{gainup}{\uparrow2.63}}$ & $65.97_{\textcolor{gainup}{\uparrow0.94}}$ & $55.73_{\textcolor{gainup}{\uparrow0.93}}$ \\
\rowcolor{rowshade}
IP-Merging & $77.40_{\textcolor{gainup}{\uparrow2.40}}$ & $46.60_{\textcolor{gainup}{\uparrow4.60}}$ & $42.76_{\textcolor{gainup}{\uparrow5.26}}$ & $66.87_{\textcolor{gainup}{\uparrow1.84}}$ & $56.67_{\textcolor{gainup}{\uparrow1.87}}$ \\
FRISM$^{\dagger}$ & $78.30_{\textcolor{gainup}{\uparrow3.30}}$ & $49.20_{\textcolor{gainup}{\uparrow7.20}}$ & $45.07_{\textcolor{gainup}{\uparrow7.57}}$ & $68.06_{\textcolor{gainup}{\uparrow3.03}}$ & $57.73_{\textcolor{gainup}{\uparrow2.93}}$ \\
\rowcolor{rowshade}
STAR & $79.10_{\textcolor{gainup}{\uparrow4.10}}$ & $49.80_{\textcolor{gainup}{\uparrow7.80}}$ & $47.70_{\textcolor{gainup}{\uparrow10.20}}$ & $68.62_{\textcolor{gainup}{\uparrow3.59}}$ & $58.27_{\textcolor{gainup}{\uparrow3.47}}$ \\
AdaRank$^{\dagger}$ & $\underline{79.70}_{\textcolor{gainup}{\uparrow4.70}}$ & $\underline{50.60}_{\textcolor{gainup}{\uparrow8.60}}$ & $\underline{48.36}_{\textcolor{gainup}{\uparrow10.86}}$ & $\underline{69.12}_{\textcolor{gainup}{\uparrow4.09}}$ & $\mathbf{60.07}_{\textcolor{gainup}{\uparrow5.27}}$ \\
\rowcolor{oursfill}
\textbf{\ourmethod{}} & $\mathbf{81.00}_{\textcolor{gainup}{\uparrow6.00}}$ & $\mathbf{52.00}_{\textcolor{gainup}{\uparrow10.00}}$ & $\mathbf{49.01}_{\textcolor{gainup}{\uparrow11.51}}$ & $\mathbf{70.06}_{\textcolor{gainup}{\uparrow5.03}}$ & $\underline{59.67}_{\textcolor{gainup}{\uparrow4.87}}$ \\
\midrule
\rowcolor{headerfill}\multicolumn{6}{c}{\textbf{H2: LLaVA-NeXT-LLaMA3-8B}} \\
Receiver & $37.40$ & $20.10$ & $13.82$ & $22.69$ & $43.80$ \\
\rowcolor{rowshade}
Full interpolation & $\underline{39.10}_{\textcolor{gainup}{\uparrow1.70}}$ & $\underline{22.40}_{\textcolor{gainup}{\uparrow2.30}}$ & $\underline{15.13}_{\textcolor{gainup}{\uparrow1.31}}$ & $\underline{24.07}_{\textcolor{gainup}{\uparrow1.38}}$ & $\mathbf{44.33}_{\textcolor{gainup}{\uparrow0.53}}$ \\
\rowcolor{oursfill}
\textbf{\ourmethod{}} & $\mathbf{42.20}_{\textcolor{gainup}{\uparrow4.80}}$ & $\mathbf{25.80}_{\textcolor{gainup}{\uparrow5.70}}$ & $\mathbf{18.09}_{\textcolor{gainup}{\uparrow4.27}}$ & $\mathbf{26.45}_{\textcolor{gainup}{\uparrow3.76}}$ & $\underline{44.13}_{\textcolor{gainup}{\uparrow0.33}}$ \\
\midrule
\rowcolor{headerfill}\multicolumn{6}{c}{\textbf{H3: Idefics2-8B}} \\
Receiver & $51.80$ & $19.40$ & $17.11$ & $21.80$ & $\mathbf{49.53}$ \\
\rowcolor{rowshade}
Full interpolation & $\underline{52.60}_{\textcolor{gainup}{\uparrow0.80}}$ & $\underline{20.40}_{\textcolor{gainup}{\uparrow1.00}}$ & $\mathbf{17.76}_{\textcolor{gainup}{\uparrow0.65}}$ & $\underline{22.55}_{\textcolor{gainup}{\uparrow0.75}}$ & $\underline{49.47}_{\textcolor{gaindown}{\downarrow0.06}}$ \\
\rowcolor{oursfill}
\textbf{\ourmethod{}} & $\mathbf{54.00}_{\textcolor{gainup}{\uparrow2.20}}$ & $\mathbf{22.20}_{\textcolor{gainup}{\uparrow2.80}}$ & $\underline{17.43}_{\textcolor{gainup}{\uparrow0.32}}$ & $\mathbf{24.15}_{\textcolor{gainup}{\uparrow2.35}}$ & $49.40_{\textcolor{gaindown}{\downarrow0.13}}$ \\
\midrule
\rowcolor{headerfill}\multicolumn{6}{c}{\textbf{H1: paired Full--Selective outcomes}} \\
Task & $N$ & Full correct & Selective correct & Gain (pp) & 95\% CI \\
\midrule
MathVista & 1,000 & 767 & 810 & \gain{4.30} & [2.14, 6.46] \\
\rowcolor{rowshade}
MathVerse & 1,000 & 453 & 520 & \gain{6.70} & [4.31, 9.09] \\
MathVision & 304 & 123 & 149 & \gain{8.55} & [3.82, 13.29] \\
\rowcolor{rowshade}
DynaMath & 5,010 & 3,310 & 3,510 & \gain{3.99} & [2.10, 5.86] \\
MMStar & 1,500 & 839 & 895 & \gain{3.73} & [2.23, 5.23] \\
\bottomrule
\end{tabular}
\endgroup
\par\vspace{2pt}
\begin{minipage}{\linewidth}\footnotesize $\dagger$: target-side adaptation. H1/H2/H3 donor steps: 120/100/110; $\lambda=0.05$. Paired gains use integer counts; DynaMath intervals resample seed questions (Appendix~\ref{sec:paired-outcomes}).\end{minipage}
\end{table}

%% file: tables/T3.tex
\begin{table}[t]
\centering
\caption{\textbf{Composition, direction controls, and generation behavior on H1.} Panels (a,b): four tasks with complete controls; five-task results: Table~\ref{tab:extended_controls}. Panel (c): 1,000 MathVista examples.}
\label{tab:T3}
\begingroup
\fontsize{8.0}{10.0}\selectfont
\setlength{\tabcolsep}{6.05pt}
\renewcommand{\arraystretch}{1.10}
\begin{tabular}{lrrrrrr}
\toprule
\rowcolor{headerfill}
Construction & $\alpha$ & $\beta$ & \shortstack{Math\\Vista} & \shortstack{Math\\Verse} & DynaMath & MMStar \\
\midrule
\rowcolor{headerfill}\multicolumn{7}{c}{\textbf{(a) Composition: initialization displacement and RL update}} \\
Receiver & 0.00 & 0.00 & 75.00 & 42.00 & 65.03 & 54.80 \\
\rowcolor{rowshade}
Initialization only & 0.05 & 0.00 & 74.70 & 41.50 & 64.65 & 54.07 \\
Direct full & 0.00 & 0.05 & 76.10 & 44.50 & 65.67 & 54.67 \\
\rowcolor{rowshade}
Direct rank-1 & 0.00 & 0.05 & 79.20 & 50.20 & 69.06 & 58.80 \\
Interpolation full & 0.05 & 0.05 & 76.70 & 45.30 & 66.07 & 55.93 \\
\rowcolor{oursfill}
Interpolation rank-1 & 0.05 & 0.05 & 81.00 & 52.00 & 70.06 & 59.67 \\
\midrule
\rowcolor{headerfill}\multicolumn{7}{c}{\textbf{(b) Direction and scale: fixed initialization displacement}} \\
Leading, unscaled & 0.05 & 0.05 & 75.20 & 42.60 & 65.27 & 54.80 \\
\rowcolor{rowshade}
Leading, tuned $\beta$ & 0.05 & 0.50 & 78.50 & 47.90 & 67.52 & 56.87 \\
Full, matrixwise gains & 0.05 & 0.05 & 77.60 & 46.50 & 66.93 & 56.33 \\
\rowcolor{rowshade}
Random, seed 42 & 0.05 & 0.05 & 73.70 & 40.50 & 63.67 & 52.93 \\
Random, 10 seeds & 0.05 & 0.05 & $74.05\pm0.45$ & $40.92\pm0.51$ & $64.11\pm0.35$ & $53.07\pm0.33$ \\
\rowcolor{rowshade}
Trailing, matched & 0.05 & 0.05 & 73.10 & 39.60 & 62.87 & 52.33 \\
\rowcolor{oursfill}
Leading rank-1 matched & 0.05 & 0.05 & 81.00 & 52.00 & 70.06 & 59.67 \\
\midrule
\rowcolor{headerfill}\multicolumn{7}{c}{\textbf{(c) H1 MathVista: output lengths and generation events}} \\
\multicolumn{2}{l}{Method} & Mean tokens & P95 tokens & Repetition & Extract fail & At cap \\
\midrule
\multicolumn{2}{l}{Receiver} & 460 & 1,550 & 8 & 5 & 6 \\
\rowcolor{rowshade}
\multicolumn{2}{l}{Full} & 575 & 2,200 & 15 & 8 & 14 \\
\rowcolor{oursfill}
\multicolumn{2}{l}{\ourmethod{}} & 430 & 1,350 & 3 & 2 & 3 \\
\bottomrule
\end{tabular}
\endgroup
\par\vspace{2pt}
\begin{minipage}{\linewidth}\footnotesize Random, 10 seeds: mean $\pm$ sample SD. Full matrixwise gains use $\rho_{j,1}^{-1/2}\Delta_j$; leading matched uses $\widehat\Delta_{j,1}$. Unscaled $\beta$ is validation-selected. Generation events can overlap.\end{minipage}
\end{table}

%% file: sections/case_study.tex
\subsubsection{Qualitative Cases and Evaluation Boundaries}
\label{sec:case-study}

\begin{figure}[H]
\centering
\setlength{\abovecaptionskip}{4pt}
\setlength{\belowcaptionskip}{0pt}

\includegraphics[width=0.90\linewidth]{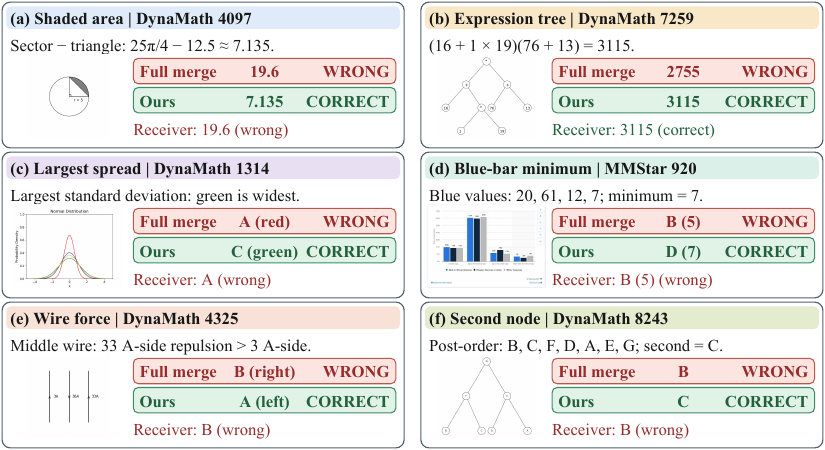}

\caption{\textbf{Six Full errors corrected by Selective-RL.}
Receiver also fails except in (b).
Cases are outcome-selected; (f) uses left-to-right post-order.
Records were retained during qualitative analysis.}
\label{fig:case-study}
\vspace{-4pt}
\end{figure}
\noindent\textbf{Observation 7: Selection can correct a shared error or avoid a merge regression.}
Figure~\ref{fig:case-study} shows six corrected Full errors: five shared failures and a merge regression in (b), where Receiver already answers $3115$. Attribution to a singular component's visual skill requires separate analysis.

These outcome-selected Qwen-family cases serve qualitative analysis; benchmark performance is evaluated separately. Appendix~\ref{sec:case-audit} reports case provenance and further response comparisons, complementing the quantitative audit.

%% file: sections/discussion.tex
\section{Conclusion}
\label{sec:conclusion}
In this paper, we propose \ourmethod{}, a selective transfer algorithm that adapts RL-stage updates for visual reasoning. The method isolates the donor's RL update, selects each eligible matrix's leading spectral direction, and restores its magnitude through Frobenius matching before composing it with a recipient VLM. This construction separates direction selection, update strength, and initialization displacement without recipient gradient training or additional inference modules. Experiments across visual reasoning benchmarks and VLM backbones show effective selective transfer; controlled ablations characterize the roles of direction, normalization, and parameter support.

%% file: sections/statements.tex
\section*{AI Use Statement}

In this work, we used generative AI tools for limited assistance with technical tasks. We have not used generative AI tools for other applicable tasks with required disclosure, and the remaining required-disclosure tasks are not applicable to this work. Additionally, we used generative AI tools to edit the manuscript to improve readability. We have reviewed all AI-assisted work. All AI-generated suggestions were independently reviewed, revised, and validated by the authors before inclusion in the final work. We take responsibility for the final content of this work, including text, claims, or artifacts produced with the aid of generative AI.

\section*{Ethics Statement}

This work uses existing model checkpoints and standard research benchmarks and does not involve the collection of private data or human-subject experiments. We follow the applicable licenses and usage conditions of the models, datasets, and software used in this study.

\section*{Reproducibility Statement}

We provide the methodological details, experimental settings, evaluation protocols, and hyperparameters needed to understand and reproduce the reported experiments in the main paper and appendices. An anonymous code repository with the corresponding implementation and supporting materials is also provided.

%% file: sections/appendix.tex
\section{Model Configurations and Evaluation Protocol}
\label{sec:configuration}
\input{sections/algorithm}
\input{tables/config}
\paragraph{Checkpoint identities.}
The receiver checkpoints and donor configurations used in the primary experiments are as follows:
\begin{itemize}[nosep,leftmargin=1.2em]
\raggedright
\item \textbf{H1:} receiver: \nolinkurl{Qwen/Qwen3-VL-8B-Instruct}; pre-RL initialization $B$: \mbox{Qwen3-8B-Base}~\citep{qwen3}; donor run: \nolinkurl{qwen3_dapo_math17k_s42_step120}.
\item \textbf{H2:} receiver: \nolinkurl{lmms-lab/llama3-llava-next-8b}; pre-RL initialization $B$: \mbox{Meta-Llama-3-8B-Instruct}; donor run: \nolinkurl{llama3_dapo_math17k_s42_step100}.
\item \textbf{H3:} receiver: \nolinkurl{HuggingFaceM4/idefics2-8b}; pre-RL initialization $B$: \mbox{Mistral-7B-v0.1}; donor run: \nolinkurl{mistral7b_dapo_math17k_s42_step110}.
\end{itemize}
Each donor is obtained from the corresponding pre-RL initialization through DAPO-style post-training. Checkpoints are saved every ten training steps within a 150-step budget, and the checkpoint used for each primary configuration is selected using source-side validation. The donor run identifiers denote the training runs used in the reported experiments.

\paragraph{Parameter support.}
The seven projections are \texttt{q/k/v/o\_proj} and \texttt{gate/up/down\_proj}. There are 36 language layers in H1 and 32 in H2/H3, giving 252/224/224 eligible matrices. Correspondence is by layer, module, and orientation, not a shared full parameter-name string. The recipient configuration and tensors outside this support remain fixed. Visual modules, embeddings, output heads, LayerNorm parameters, and biases are excluded, also avoiding transfer of incompatible multimodal vocabulary extensions.
The anonymous repository includes inference and evaluation support based on VLMEvalKit~\citep{vlmevalkit}.
\input{tables/benchmarks}
\input{tables/validation}

\section{Extended Method and Control Comparisons}
\label{sec:extended-comparisons}
Table~\ref{tab:extended_main} retains every reported baseline and available five-task score. The main comparison presents rows with all five measurements available; this appendix preserves the broader method coverage. Table~\ref{tab:extended_controls} reports the five-task controls, including MathVision, while the main control block uses four tasks with complete coverage. --- indicates an unavailable measurement.
\input{tables/extended_main}
\input{tables/extended_controls}

\section{Additional Mechanism and Spectral Analyses}
\label{sec:additional-controls}
Tables~\ref{tab:T4}--\ref{tab:T6} expand the rank, coefficient, and support comparisons. Tables~\ref{tab:rho_group}--\ref{tab:rho_layer} report retained-energy distributions without fitting an association across configurations.
\input{tables/T4}
\input{tables/T5}
\input{tables/T6}
\input{tables/rho_group}
\input{tables/rho_module}
\input{tables/rho_layer}
\input{tables/normalization}
\input{tables/stage_isolation}
\input{tables/beta_sweep}
\input{tables/scaling}
\input{tables/random_seed}

\section{Source Recovery, Donor Variation, and Boundaries}
\label{sec:recorded-evidence}
Accuracy retention is $A_{\mathrm{recon}}/A_R$ and gain retention is $(A_{\mathrm{recon}}-A_B)/(A_R-A_B)$, each expressed as a percentage. These describe recovery of donor behavior; recipient transfer is evaluated separately. Donor stages belong to one trajectory, while the seed comparison repeats donor training.
\input{tables/T2}
\input{tables/T9}
\input{tables/T8}
\input{tables/donor_seed}

\section{Paired Outcomes and Variant Robustness}
\label{sec:paired-outcomes}
Let $(a,b,c,d)$ denote both-correct, Full-only, Selective-only, and both-wrong counts. With $n=a+b+c+d$ and $\widehat\delta=(c-b)/n$, the paired normal interval for independent evaluation examples is
\begin{equation}
 100\left[\widehat\delta\ \pm\ 1.96\sqrt{\frac{(b+c)/n-\widehat\delta^2}{n-1}}\right].
 \label{eq:paired-ci}
\end{equation}
For DynaMath the resampling unit is the seed question. If $y_{s,v}^{M}\in\{0,1\}$ is correctness for method $M$ on variant $v$ of seed $s$, then
\begin{equation}
 d_s=\frac{1}{10}\sum_{v=1}^{10}(y_{s,v}^{\mathrm{Selective}}-y_{s,v}^{\mathrm{Full}}),
 \qquad \widehat\delta=\frac{1}{501}\sum_{s=1}^{501}d_s.
 \label{eq:cluster-difference}
\end{equation}
Seed-cluster bootstrap resamples the 501 questions with replacement, retaining the ten individual variant outcomes within each question. It does not impose equal outcomes within a question. The reported DynaMath interval is $[2.10,5.86]$ percentage points. The paired counts yield $3{,}310$ Full-correct and $3{,}510$ Selective-correct variants; $113+313=426$ variants have different correctness outcomes.
\input{tables/T13}
\input{tables/dyna_seed}

\section{Generation Diagnostics and Resource Accounting}
\label{sec:resources}
Output diagnostics are measured on H1 MathVista. Repetition, extraction failure, and reaching the generation cap may overlap. The separate analysis of Selective-only wins uses the mutually exclusive categories in Table~\ref{tab:error_types}. The evaluator audit compares extracted labels with final-answer correctness on 200 manually inspected examples per model, separately from full benchmark scores.
\input{tables/T11}
\input{tables/error_types}
\input{tables/evaluator}
\input{tables/T10}
\input{tables/T12}
\input{tables/T14}
Construction totals sum decomposition, other processing, writing, and calibration. Validation search is additional and reported in GPU-hours; donor training is a separate upstream cost. Parameter count and VRAM characterize the dense merged model. Timing and VRAM measurements use the same hardware/software environment. Output length measures token use; inference latency is a separate timing metric.

%% file: sections/algorithm.tex
\begingroup
\definecolor{srAlgTeal}{HTML}{2B7A78}
\definecolor{srAlgBlue}{HTML}{345F8F}
\definecolor{srAlgPurple}{HTML}{76568D}
\newcommand{\srAlgComment}[2]{\hspace{0.5em}\hfill\mbox{\color{#1}\fontsize{9}{11}\selectfont\ttfamily\bfseries// #2}}
\begin{algorithm}[H]
\caption{\textbf{Fixed Matrixwise Rank-One Transfer}}
\label{alg:selective-transfer}
\centering
{\fontsize{9}{11}\selectfont
\begin{tabularx}{\linewidth}{@{}r@{\hspace{0.65em}}X@{}}
1 & \textbf{Input:} recipient $V$, actual pre-RL checkpoint $B$, donor $R$, matrix support $\mathcal S$, coefficients $\alpha,\beta$. \\
2 & Verify aligned tensor coordinates and consistent support and updates for tied parameters. \\
3 & $W\gets\operatorname{copy}(V)$. \srAlgComment{srAlgBlue}{copy all tensors} \\
4 & \textbf{for each} $j\in\mathcal S$ \textbf{do} \\
5 & \hspace*{1em}$\Gamma_j\gets B_j-V_j$; $\Delta_j\gets R_j-B_j$. \srAlgComment{srAlgTeal}{isolate RL} \\
6 & \hspace*{1em}$\eta_j\gets\sqrt{\langle\Delta_j,\Delta_j\rangle_F}=\|\Delta_j\|_F$. \\
7 & \hspace*{1em}\textbf{if} $\eta_j=0$ \textbf{then} $Z_j\gets0$ \srAlgComment{srAlgPurple}{zero RL term} \\
8 & \hspace*{1em}\textbf{else} $(\sigma_{j,1},u_{j,1},v_{j,1})\gets\operatorname{svd}_1(\Delta_j)$ \\
9 & \hspace*{2em}$Z_j\gets\eta_j u_{j,1}v_{j,1}^{\top}=\mathcal N_1(\Delta_j)$. \srAlgComment{srAlgTeal}{match update norm} \\
10 & \hspace*{1em}\textbf{end if} \\
11 & \hspace*{1em}$W_j\gets V_j+\alpha\Gamma_j+\beta Z_j$. \srAlgComment{srAlgBlue}{compose both updates} \\
12 & \textbf{end for} \\
13 & \textbf{Return} $W$; $\forall j\notin\mathcal S,\ W_j=V_j$. \\
\midrule
 & \textbf{Routes:} interpolation $\alpha=\beta=\lambda$; direct addition $\alpha=0$ with coefficient $\beta$. \\
\end{tabularx}
}
\end{algorithm}
\endgroup

%% file: tables/config.tex
\begin{table}[H]
\centering
\caption{\textbf{Model groups and donor selection.} All donors use matched DAPO-style training on DAPO-Math-17K.}
\label{tab:config}
\begingroup
\fontsize{8.0}{10.0}\selectfont
\setlength{\tabcolsep}{4pt}
\renewcommand{\arraystretch}{1.12}
\begin{tabular}{@{}lrrr@{}}
\toprule
\rowcolor{headerfill}
Configuration & H1: primary & H2: cross-family & H3: cross-backbone \\
\midrule
Receiver & Qwen3-VL-8B-Instruct & LLaVA-NeXT-LLaMA3-8B & Idefics2-8B \\
\rowcolor{rowshade}
LM family & Qwen3 & LLaMA3 & Mistral \\
Layers / matrices & 36 / 252 & 32 / 224 & 32 / 224 \\
\rowcolor{rowshade}
Donor seed & 42 & 42 & 42 \\
Selected step & \textbf{120} & \textbf{100} & \textbf{110} \\
\rowcolor{rowshade}
Training budget & 150 steps & 150 steps & 150 steps \\
Save interval & 10 steps & 10 steps & 10 steps \\
\rowcolor{rowshade}
Donor selection & \multicolumn{3}{c}{Source-side validation only} \\
Coefficient & \multicolumn{3}{c}{$0.05$: H1 held-out selection, then frozen for H2/H3} \\
\bottomrule
\end{tabular}
\endgroup
\par\vspace{2pt}
\begin{minipage}{\linewidth}\footnotesize The actual initialization used by each donor defines $B$. All groups use bf16 and rank 1; visual components, embeddings, output heads, LayerNorm, and bias remain frozen.\end{minipage}
\end{table}

%% file: tables/benchmarks.tex
\begin{table}[H]
\centering
\caption{\textbf{Evaluation tasks and sample sizes.} Main results use the first five tasks; the remaining four assess broader capabilities.}
\label{tab:benchmarks}
\begingroup
\fontsize{9}{11}\selectfont
\setlength{\tabcolsep}{4pt}
\renewcommand{\arraystretch}{1.12}
\begin{tabular}{@{}lrr@{}}
\toprule
\rowcolor{headerfill}
Benchmark & $N$ & Metric \\
\midrule
MathVista & 1,000 & Accuracy (\%) \\
\rowcolor{rowshade}
MathVerse & 1,000 & Accuracy (\%) \\
MathVision & 304 & Accuracy (\%) \\
\rowcolor{rowshade}
DynaMath & 5,010 & Accuracy (\%) \\
MMStar & 1,500 & Accuracy (\%) \\
\rowcolor{rowshade}
TextVQA & 1,000 & VQA soft score \\
MMMU & 1,000 & Accuracy (\%) \\
\rowcolor{rowshade}
OCRBench & 1,000 & Score (\%) \\
POPE & 3,000 & Accuracy (\%) \\
\bottomrule
\end{tabular}
\endgroup
\end{table}

%% file: tables/validation.tex
\begin{table}[H]
\centering
\caption{\textbf{H1 held-out coefficient selection.} $\lambda=0.05$ leads both validation metrics and is then frozen for H2 and H3.}
\label{tab:validation}
\begingroup
\fontsize{9}{11}\selectfont
\setlength{\tabcolsep}{4pt}
\renewcommand{\arraystretch}{1.12}
\begin{tabular}{@{}lrr@{}}
\toprule
\rowcolor{headerfill}
$\lambda$ & MathVista-val & DynaMath-val \\
\midrule
0.01 & 76.90 & 66.10 \\
\rowcolor{rowshade}
0.025 & 79.00 & 68.35 \\
0.05 & 80.20 & 69.40 \\
\rowcolor{rowshade}
0.075 & 79.90 & 69.10 \\
0.1 & 79.50 & 68.75 \\
\bottomrule
\end{tabular}
\endgroup
\end{table}

%% file: tables/extended_main.tex
\begin{table}[H]
\centering
\caption{\textbf{Extended comparison across all reported merging methods.} Accuracy (\%). Colored subscripts show changes from the Receiver within each group; changes use displayed scores. Bold/underline: best/second-best available score per group and task.}
\label{tab:extended_main}
\begingroup
\fontsize{8.6}{10.6}\selectfont
\setlength{\tabcolsep}{4pt}
\renewcommand{\arraystretch}{1.12}
\begin{tabular}{@{}lrrrrr@{}}
\toprule
\rowcolor{headerfill}
Method & \shortstack{Math\\Vista} & \shortstack{Math\\Verse} & \shortstack{Math\\Vision} & DynaMath & MMStar \\
\midrule
\rowcolor{headerfill}\multicolumn{6}{c}{\textbf{H1: Qwen3-VL-8B-Instruct}} \\
\rowcolor{rowshade}
Receiver & $75.00$ & $42.00$ & $37.50$ & $65.03$ & $54.80$ \\
Task Arithmetic & $76.10_{\textcolor{gainup}{\uparrow1.10}}$ & $44.50_{\textcolor{gainup}{\uparrow2.50}}$ & $39.47_{\textcolor{gainup}{\uparrow1.97}}$ & $65.67_{\textcolor{gainup}{\uparrow0.64}}$ & $54.67_{\textcolor{gaindown}{\downarrow0.13}}$ \\
\rowcolor{rowshade}
Full interpolation & $76.70_{\textcolor{gainup}{\uparrow1.70}}$ & $45.30_{\textcolor{gainup}{\uparrow3.30}}$ & $40.46_{\textcolor{gainup}{\uparrow2.96}}$ & $66.07_{\textcolor{gainup}{\uparrow1.04}}$ & $55.93_{\textcolor{gainup}{\uparrow1.13}}$ \\
TIES & $76.40_{\textcolor{gainup}{\uparrow1.40}}$ & $44.80_{\textcolor{gainup}{\uparrow2.80}}$ & $39.80_{\textcolor{gainup}{\uparrow2.30}}$ & $65.77_{\textcolor{gainup}{\uparrow0.74}}$ & $55.53_{\textcolor{gainup}{\uparrow0.73}}$ \\
\rowcolor{rowshade}
DARE & $76.30_{\textcolor{gainup}{\uparrow1.30}}$ & $45.00_{\textcolor{gainup}{\uparrow3.00}}$ & $40.13_{\textcolor{gainup}{\uparrow2.63}}$ & $65.97_{\textcolor{gainup}{\uparrow0.94}}$ & $55.73_{\textcolor{gainup}{\uparrow0.93}}$ \\
TSV & $77.90_{\textcolor{gainup}{\uparrow2.90}}$ & $47.80_{\textcolor{gainup}{\uparrow5.80}}$ & --- & --- & $57.07_{\textcolor{gainup}{\uparrow2.27}}$ \\
\rowcolor{rowshade}
LoRE & $78.40_{\textcolor{gainup}{\uparrow3.40}}$ & $48.70_{\textcolor{gainup}{\uparrow6.70}}$ & --- & $67.90_{\textcolor{gainup}{\uparrow2.87}}$ & $57.47_{\textcolor{gainup}{\uparrow2.67}}$ \\
IP-Merging & $77.40_{\textcolor{gainup}{\uparrow2.40}}$ & $46.60_{\textcolor{gainup}{\uparrow4.60}}$ & $42.76_{\textcolor{gainup}{\uparrow5.26}}$ & $66.87_{\textcolor{gainup}{\uparrow1.84}}$ & $56.67_{\textcolor{gainup}{\uparrow1.87}}$ \\
\rowcolor{rowshade}
FRISM$^{\dagger}$ & $78.30_{\textcolor{gainup}{\uparrow3.30}}$ & $49.20_{\textcolor{gainup}{\uparrow7.20}}$ & $45.07_{\textcolor{gainup}{\uparrow7.57}}$ & $68.06_{\textcolor{gainup}{\uparrow3.03}}$ & $57.73_{\textcolor{gainup}{\uparrow2.93}}$ \\
STAR & $79.10_{\textcolor{gainup}{\uparrow4.10}}$ & $49.80_{\textcolor{gainup}{\uparrow7.80}}$ & $47.70_{\textcolor{gainup}{\uparrow10.20}}$ & $68.62_{\textcolor{gainup}{\uparrow3.59}}$ & $58.27_{\textcolor{gainup}{\uparrow3.47}}$ \\
\rowcolor{rowshade}
AdaRank$^{\dagger}$ & $\underline{79.70}_{\textcolor{gainup}{\uparrow4.70}}$ & $\underline{50.60}_{\textcolor{gainup}{\uparrow8.60}}$ & $\underline{48.36}_{\textcolor{gainup}{\uparrow10.86}}$ & $\underline{69.12}_{\textcolor{gainup}{\uparrow4.09}}$ & $\mathbf{60.07}_{\textcolor{gainup}{\uparrow5.27}}$ \\
\textbf{\ourmethod{}} & $\mathbf{81.00}_{\textcolor{gainup}{\uparrow6.00}}$ & $\mathbf{52.00}_{\textcolor{gainup}{\uparrow10.00}}$ & $\mathbf{49.01}_{\textcolor{gainup}{\uparrow11.51}}$ & $\mathbf{70.06}_{\textcolor{gainup}{\uparrow5.03}}$ & $\underline{59.67}_{\textcolor{gainup}{\uparrow4.87}}$ \\
\midrule
\rowcolor{headerfill}\multicolumn{6}{c}{\textbf{H2: LLaVA-NeXT-LLaMA3-8B}} \\
\rowcolor{rowshade}
Receiver & $37.40$ & $20.10$ & $13.82$ & $22.69$ & $43.80$ \\
Full interpolation & $39.10_{\textcolor{gainup}{\uparrow1.70}}$ & $22.40_{\textcolor{gainup}{\uparrow2.30}}$ & $\underline{15.13}_{\textcolor{gainup}{\uparrow1.31}}$ & $\underline{24.07}_{\textcolor{gainup}{\uparrow1.38}}$ & $44.33_{\textcolor{gainup}{\uparrow0.53}}$ \\
\rowcolor{rowshade}
STAR & $40.60_{\textcolor{gainup}{\uparrow3.20}}$ & $24.20_{\textcolor{gainup}{\uparrow4.10}}$ & --- & --- & $\underline{44.53}_{\textcolor{gainup}{\uparrow0.73}}$ \\
AdaRank$^{\dagger}$ & $\underline{41.00}_{\textcolor{gainup}{\uparrow3.60}}$ & $\underline{24.80}_{\textcolor{gainup}{\uparrow4.70}}$ & --- & --- & $\mathbf{44.93}_{\textcolor{gainup}{\uparrow1.13}}$ \\
\rowcolor{rowshade}
\textbf{\ourmethod{}} & $\mathbf{42.20}_{\textcolor{gainup}{\uparrow4.80}}$ & $\mathbf{25.80}_{\textcolor{gainup}{\uparrow5.70}}$ & $\mathbf{18.09}_{\textcolor{gainup}{\uparrow4.27}}$ & $\mathbf{26.45}_{\textcolor{gainup}{\uparrow3.76}}$ & $44.13_{\textcolor{gainup}{\uparrow0.33}}$ \\
\midrule
\rowcolor{headerfill}\multicolumn{6}{c}{\textbf{H3: Idefics2-8B}} \\
Receiver & $51.80$ & $19.40$ & $17.11$ & $21.80$ & $49.53$ \\
\rowcolor{rowshade}
Full interpolation & $52.60_{\textcolor{gainup}{\uparrow0.80}}$ & $20.40_{\textcolor{gainup}{\uparrow1.00}}$ & $\mathbf{17.76}_{\textcolor{gainup}{\uparrow0.65}}$ & $\underline{22.55}_{\textcolor{gainup}{\uparrow0.75}}$ & $49.47_{\textcolor{gaindown}{\downarrow0.06}}$ \\
STAR & $53.20_{\textcolor{gainup}{\uparrow1.40}}$ & $21.40_{\textcolor{gainup}{\uparrow2.00}}$ & --- & --- & $\underline{49.60}_{\textcolor{gainup}{\uparrow0.07}}$ \\
\rowcolor{rowshade}
AdaRank$^{\dagger}$ & $\underline{53.60}_{\textcolor{gainup}{\uparrow1.80}}$ & $\underline{21.70}_{\textcolor{gainup}{\uparrow2.30}}$ & --- & --- & $\mathbf{50.00}_{\textcolor{gainup}{\uparrow0.47}}$ \\
\textbf{\ourmethod{}} & $\mathbf{54.00}_{\textcolor{gainup}{\uparrow2.20}}$ & $\mathbf{22.20}_{\textcolor{gainup}{\uparrow2.80}}$ & $\underline{17.43}_{\textcolor{gainup}{\uparrow0.32}}$ & $\mathbf{24.15}_{\textcolor{gainup}{\uparrow2.35}}$ & $49.40_{\textcolor{gaindown}{\downarrow0.13}}$ \\
\bottomrule
\end{tabular}
\endgroup
\par\vspace{2pt}
\begin{minipage}{\linewidth}\footnotesize $\dagger$: target-side adaptation. --- indicates an unavailable measurement. H1/H2/H3 use donor checkpoints 120/100/110, respectively, with $\lambda=0.05$.\end{minipage}
\end{table}

%% file: tables/extended_controls.tex
\begin{table}[H]
\centering
\caption{\textbf{Five-benchmark composition and direction controls.} The same five benchmarks are used across controls. The ten-seed random row reports mean $\pm$ sample standard deviation.}
\label{tab:extended_controls}
\begingroup
\fontsize{8.1}{10.1}\selectfont
\setlength{\tabcolsep}{4pt}
\renewcommand{\arraystretch}{1.12}
\begin{tabular}{@{}lrrrrrrr@{}}
\toprule
\rowcolor{headerfill}
Construction & $\alpha$ & $\beta$ & \shortstack{Math\\Vista} & \shortstack{Math\\Verse} & \shortstack{Math\\Vision} & DynaMath & MMStar \\
\midrule
\rowcolor{headerfill}\multicolumn{8}{c}{\textbf{(a) Composition: initialization displacement and RL update}} \\
\rowcolor{rowshade}
Receiver & 0.00 & 0.00 & 75.00 & 42.00 & 37.50 & 65.03 & 54.80 \\
Initialization only & 0.05 & 0.00 & 74.70 & 41.50 & 37.17 & 64.65 & 54.07 \\
\rowcolor{rowshade}
Direct full & 0.00 & 0.05 & 76.10 & 44.50 & 39.47 & 65.67 & 54.67 \\
Direct rank-1 & 0.00 & 0.05 & 79.20 & 50.20 & 47.04 & 69.06 & 58.80 \\
\rowcolor{rowshade}
Interpolation full & 0.05 & 0.05 & 76.70 & 45.30 & 40.46 & 66.07 & 55.93 \\
\rowcolor{oursfill}
Interpolation rank-1 & 0.05 & 0.05 & 81.00 & 52.00 & 49.01 & 70.06 & 59.67 \\
\midrule
\rowcolor{headerfill}\multicolumn{8}{c}{\textbf{(b) Direction and scale: fixed initialization displacement}} \\
\rowcolor{rowshade}
Leading, unscaled & 0.05 & 0.05 & 75.20 & 42.60 & 38.16 & 65.27 & 54.80 \\
Leading, tuned $\beta$ & 0.05 & 0.50 & 78.50 & 47.90 & --- & 67.52 & 56.87 \\
\rowcolor{rowshade}
Full, matrixwise gains & 0.05 & 0.05 & 77.60 & 46.50 & --- & 66.93 & 56.33 \\
Random, seed 42 & 0.05 & 0.05 & 73.70 & 40.50 & 35.86 & 63.67 & 52.93 \\
\rowcolor{rowshade}
Random, 10 seeds & 0.05 & 0.05 & $74.05\pm0.45$ & $40.92\pm0.51$ & --- & $64.11\pm0.35$ & $53.07\pm0.33$ \\
Trailing, matched & 0.05 & 0.05 & 73.10 & 39.60 & 34.54 & 62.87 & 52.33 \\
\rowcolor{oursfill}
Leading rank-1 matched & 0.05 & 0.05 & 81.00 & 52.00 & 49.01 & 70.06 & 59.67 \\
\bottomrule
\end{tabular}
\endgroup
\par\vspace{2pt}
\begin{minipage}{\linewidth}\footnotesize Full matrixwise gains use $\rho_{j,1}^{-1/2}\Delta_j$; leading matched uses $\widehat\Delta_{j,1}$. The unscaled coefficient is chosen on validation. --- indicates an unavailable measurement.\end{minipage}
\end{table}

%% file: tables/T4.tex
\begin{table}[H]
\centering
\caption{\textbf{Rank sensitivity on H1.} Frobenius matching is applied at every finite rank.}
\label{tab:T4}
\begingroup
\fontsize{9}{11}\selectfont
\setlength{\tabcolsep}{4pt}
\renewcommand{\arraystretch}{1.12}
\begin{tabular}{@{}lrrrrr@{}}
\toprule
\rowcolor{headerfill}
Rank & \shortstack{Math\\Vista} & \shortstack{Math\\Verse} & \shortstack{Math\\Vision} & DynaMath & MMStar \\
\midrule
1 & 81.00 & 52.00 & 49.01 & 70.06 & 59.67 \\
\rowcolor{rowshade}
2 & 80.70 & 51.60 & --- & 69.66 & 59.27 \\
4 & 80.30 & 50.80 & 47.70 & 69.06 & 58.67 \\
\rowcolor{rowshade}
8 & 79.50 & 49.40 & --- & 68.26 & 58.00 \\
16 & 78.30 & 47.60 & --- & 67.27 & 57.00 \\
\rowcolor{rowshade}
Full & 76.70 & 45.30 & 40.46 & 66.07 & 55.93 \\
\bottomrule
\end{tabular}
\endgroup
\par\vspace{2pt}
\begin{minipage}{\linewidth}\footnotesize Full retains the complete update. --- indicates an unavailable measurement.\end{minipage}
\end{table}

%% file: tables/T5.tex
\begin{table}[H]
\centering
\caption{\textbf{Complete test-set coefficient sweep.} The main coefficient $0.05$ is selected using H1 validation, not this test sweep.}
\label{tab:T5}
\begingroup
\fontsize{8.5}{10.5}\selectfont
\setlength{\tabcolsep}{4pt}
\renewcommand{\arraystretch}{1.12}
\begin{tabular}{@{}lrrrrrr@{}}
\toprule
\rowcolor{headerfill}
$\lambda$ & \shortstack{MathVista\\Full} & \shortstack{MathVista\\Rank-1} & \shortstack{DynaMath\\Full} & \shortstack{DynaMath\\Rank-1} & \shortstack{MMStar\\Full} & \shortstack{MMStar\\Rank-1} \\
\midrule
0 & 75.00 & 75.00 & 65.03 & 65.03 & 54.80 & 54.80 \\
\rowcolor{rowshade}
0.01 & 75.40 & 77.40 & 65.27 & 66.67 & 55.33 & 56.60 \\
0.025 & 76.20 & 79.90 & 65.71 & 69.26 & 55.67 & 58.60 \\
\rowcolor{rowshade}
0.05 & 76.70 & 81.00 & 66.07 & 70.06 & 55.93 & 59.67 \\
0.1 & 76.00 & 80.50 & 65.47 & 69.66 & 54.93 & 59.27 \\
\rowcolor{rowshade}
0.2 & 74.40 & 78.90 & 63.87 & 68.26 & 53.00 & 57.73 \\
0.5 & 70.00 & 76.50 & 59.88 & 66.07 & 49.00 & 55.67 \\
\rowcolor{rowshade}
1 & 61.00 & 75.20 & 52.89 & 65.07 & 42.33 & 55.00 \\
\bottomrule
\end{tabular}
\endgroup
\end{table}

%% file: tables/T6.tex
\begin{table}[H]
\centering
\caption{\textbf{Eligible support on H1.} The remaining recipient parameters stay fixed.}
\label{tab:T6}
\begingroup
\fontsize{8.5}{10.5}\selectfont
\setlength{\tabcolsep}{4pt}
\renewcommand{\arraystretch}{1.12}
\begin{tabular}{@{}lrrrrrr@{}}
\toprule
\rowcolor{headerfill}
Support & Matrices & \shortstack{Math\\Vista} & \shortstack{Math\\Verse} & \shortstack{Math\\Vision} & DynaMath & MMStar \\
\midrule
QKV only & 108 & 78.10 & 47.60 & --- & --- & 57.47 \\
\rowcolor{rowshade}
Attention only & 144 & 79.40 & 49.30 & --- & 68.52 & 58.47 \\
MLP only & 108 & 78.80 & 48.60 & --- & 68.88 & 57.93 \\
\rowcolor{rowshade}
O + Down only & 72 & 77.90 & 46.90 & --- & 67.54 & 57.20 \\
Attention + MLP & 252 & 81.00 & 52.00 & 49.01 & 70.06 & 59.67 \\
\bottomrule
\end{tabular}
\endgroup
\par\vspace{2pt}
\begin{minipage}{\linewidth}\footnotesize QKV: query/key/value projections; O: attention output; Down: MLP down projection. --- indicates an unavailable measurement.\end{minipage}
\end{table}

%% file: tables/rho_group.tex
\begin{table}[H]
\centering
\caption{\textbf{Spectral concentration by model configuration.} $\rho_1$ is retained squared Frobenius energy; amplification is $\rho_1^{-1/2}$.}
\label{tab:rho_group}
\begingroup
\fontsize{8.4}{10.4}\selectfont
\setlength{\tabcolsep}{4pt}
\renewcommand{\arraystretch}{1.12}
\begin{tabular}{@{}lrrrrrr@{}}
\toprule
\rowcolor{headerfill}
Configuration & \shortstack{Attention\\median} & \shortstack{MLP\\median} & Overall & Q1 & Q3 & \shortstack{Median\\amplification} \\
\midrule
H1 Qwen RL & 0.24 & 0.15 & 0.18 & 0.11 & 0.31 & 2.36 \\
\rowcolor{rowshade}
H2 LLaMA RL & 0.17 & 0.11 & 0.13 & 0.08 & 0.24 & 2.77 \\
H3 Mistral RL & 0.12 & 0.08 & 0.09 & 0.05 & 0.17 & 3.33 \\
\bottomrule
\end{tabular}
\endgroup
\end{table}

%% file: tables/rho_module.tex
\begin{table}[H]
\centering
\caption{\textbf{H1 spectral energy by module.} Quartiles summarize the distribution across eligible matrices.}
\label{tab:rho_module}
\begingroup
\fontsize{9}{11}\selectfont
\setlength{\tabcolsep}{4pt}
\renewcommand{\arraystretch}{1.12}
\begin{tabular}{@{}lrrrr@{}}
\toprule
\rowcolor{headerfill}
Module & Mean $\rho_1$ & Median $\rho_1$ & Q1 & Q3 \\
\midrule
q\_proj & 0.27 & 0.25 & 0.17 & 0.35 \\
\rowcolor{rowshade}
k\_proj & 0.24 & 0.22 & 0.15 & 0.31 \\
v\_proj & 0.25 & 0.23 & 0.16 & 0.33 \\
\rowcolor{rowshade}
o\_proj & 0.21 & 0.19 & 0.12 & 0.28 \\
gate\_proj & 0.16 & 0.14 & 0.09 & 0.22 \\
\rowcolor{rowshade}
up\_proj & 0.15 & 0.13 & 0.08 & 0.21 \\
down\_proj & 0.18 & 0.16 & 0.10 & 0.25 \\
\bottomrule
\end{tabular}
\endgroup
\end{table}

%% file: tables/rho_layer.tex
\begin{table}[H]
\centering
\caption{\textbf{H1 spectral energy by depth.}}
\label{tab:rho_layer}
\begingroup
\fontsize{9}{11}\selectfont
\setlength{\tabcolsep}{4pt}
\renewcommand{\arraystretch}{1.12}
\begin{tabular}{@{}lrr@{}}
\toprule
\rowcolor{headerfill}
Layer region & Median $\rho_1$ & Median amplification \\
\midrule
Early 1-12 & 0.13 & 2.77 \\
\rowcolor{rowshade}
Middle 13-24 & 0.22 & 2.13 \\
Late 25-36 & 0.19 & 2.29 \\
\bottomrule
\end{tabular}
\endgroup
\end{table}

%% file: tables/normalization.tex
\begin{table}[H]
\centering
\caption{\textbf{Normalization ablation on H1.} Leading rank-one updates are compared at the same composition coefficients.}
\label{tab:normalization}
\begingroup
\fontsize{9}{11}\selectfont
\setlength{\tabcolsep}{4pt}
\renewcommand{\arraystretch}{1.12}
\begin{tabular}{@{}lrrrrr@{}}
\toprule
\rowcolor{headerfill}
Normalization & \shortstack{Math\\Vista} & \shortstack{Math\\Verse} & \shortstack{Math\\Vision} & DynaMath & MMStar \\
\midrule
None & 75.20 & 42.60 & 38.16 & 65.27 & 54.80 \\
\rowcolor{rowshade}
Nuclear-norm match & 80.00 & 50.10 & --- & 69.16 & 58.67 \\
Frobenius match & 81.00 & 52.00 & 49.01 & 70.06 & 59.67 \\
\bottomrule
\end{tabular}
\endgroup
\par\vspace{2pt}
\begin{minipage}{\linewidth}\footnotesize Frobenius and nuclear matching use the corresponding norm of the full matrix as the target. --- indicates an unavailable measurement.\end{minipage}
\end{table}

%% file: tables/stage_isolation.tex
\begin{table}[H]
\centering
\caption{\textbf{Stage-reference ablation.} The actual donor initialization is compared with an earlier reference checkpoint.}
\label{tab:stage_isolation}
\begingroup
\fontsize{8.7}{10.7}\selectfont
\setlength{\tabcolsep}{4pt}
\renewcommand{\arraystretch}{1.12}
\begin{tabular}{@{}lrrrrr@{}}
\toprule
\rowcolor{headerfill}
Reference control & \shortstack{Math\\Vista} & \shortstack{Math\\Verse} & \shortstack{Math\\Vision} & DynaMath & MMStar \\
\midrule
Full, actual pre-RL & 76.70 & 45.30 & 40.46 & 66.07 & 55.93 \\
\rowcolor{rowshade}
Rank-1, actual pre-RL & 81.00 & 52.00 & 49.01 & 70.06 & 59.67 \\
Full, earlier reference & 75.60 & 43.80 & --- & 65.41 & 54.93 \\
\rowcolor{rowshade}
Rank-1, earlier reference & 77.80 & 46.90 & --- & 67.11 & 56.40 \\
\bottomrule
\end{tabular}
\endgroup
\par\vspace{2pt}
\begin{minipage}{\linewidth}\footnotesize All variants use $\alpha=\beta=0.05$. --- indicates an unavailable measurement.\end{minipage}
\end{table}

%% file: tables/beta_sweep.tex
\begin{table}[H]
\centering
\caption{\textbf{Unscaled leading rank-one coefficient sweep.} Initialization displacement remains fixed at $\alpha=0.05$.}
\label{tab:beta_sweep}
\begingroup
\fontsize{9}{11}\selectfont
\setlength{\tabcolsep}{4pt}
\renewcommand{\arraystretch}{1.12}
\begin{tabular}{@{}lrrr@{}}
\toprule
\rowcolor{headerfill}
$\beta$ & MathVista & DynaMath & MMStar \\
\midrule
0.05 & 75.20 & 65.27 & 54.80 \\
\rowcolor{rowshade}
0.1 & 76.40 & 65.87 & 55.47 \\
0.2 & 77.60 & 66.91 & 56.33 \\
\rowcolor{rowshade}
0.5 & 78.50 & 67.52 & 56.87 \\
1 & 77.80 & 66.95 & 56.40 \\
\rowcolor{rowshade}
2 & 74.90 & 63.95 & 53.87 \\
\bottomrule
\end{tabular}
\endgroup
\end{table}

%% file: tables/scaling.tex
\begin{table}[H]
\centering
\caption{\textbf{Separating direction selection from matrixwise amplification.}}
\label{tab:scaling}
\begingroup
\fontsize{8.8}{10.8}\selectfont
\setlength{\tabcolsep}{4pt}
\renewcommand{\arraystretch}{1.12}
\begin{tabular}{@{}lrrr@{}}
\toprule
\rowcolor{headerfill}
Control & MathVista & DynaMath & MMStar \\
\midrule
Full, global beta & 76.70 & 66.07 & 55.93 \\
\rowcolor{rowshade}
Full, rank-1 matrixwise gains & 77.60 & 66.93 & 56.33 \\
Rank-1, tuned global beta & 78.50 & 67.52 & 56.87 \\
\rowcolor{rowshade}
Rank-1, Frobenius matched & 81.00 & 70.06 & 59.67 \\
\bottomrule
\end{tabular}
\endgroup
\end{table}

%% file: tables/random_seed.tex
\begin{table}[H]
\centering
\caption{\textbf{Random-direction robustness.} Ten matched rank-one direction draws at $\alpha=\beta=0.05$.}
\label{tab:random_seed}
\begingroup
\fontsize{9}{11}\selectfont
\setlength{\tabcolsep}{4pt}
\renewcommand{\arraystretch}{1.12}
\begin{tabular}{@{}lrrr@{}}
\toprule
\rowcolor{headerfill}
Direction seed & MathVista & DynaMath & MMStar \\
\midrule
0 & 74.60 & 64.43 & 53.27 \\
\rowcolor{rowshade}
1 & 73.80 & 63.87 & 52.87 \\
2 & 74.10 & 64.07 & 53.13 \\
\rowcolor{rowshade}
3 & 73.40 & 63.55 & 52.47 \\
4 & 74.70 & 64.51 & 53.60 \\
\rowcolor{rowshade}
5 & 73.90 & 63.91 & 52.93 \\
6 & 74.30 & 64.23 & 53.20 \\
\rowcolor{rowshade}
7 & 73.50 & 63.67 & 52.73 \\
8 & 74.40 & 64.35 & 53.40 \\
\rowcolor{rowshade}
9 & 73.80 & 64.49 & 53.07 \\
\textbf{Mean $\pm$ SD} & $74.05\pm0.45$ & $64.11\pm0.35$ & $53.07\pm0.33$ \\
\bottomrule
\end{tabular}
\endgroup
\par\vspace{2pt}
\begin{minipage}{\linewidth}\footnotesize Summary statistics are recomputed from the displayed ten runs using sample SD.\end{minipage}
\end{table}

%% file: tables/T2.tex
\begin{table}[H]
\centering
\caption{\textbf{Source-side reconstruction.} MATH~\citep{mathdataset} and GSM8K~\citep{gsm8k} accuracy; retention ratios use the donor accuracy or its gain over the pre-RL model.}
\label{tab:T2}
\begingroup
\fontsize{9}{11}\selectfont
\setlength{\tabcolsep}{4pt}
\renewcommand{\arraystretch}{1.12}
\begin{tabular}{@{}lrrrr@{}}
\toprule
\rowcolor{headerfill}
Reconstruction & MATH & GSM8K & \shortstack{Accuracy\\retention (\%)} & \shortstack{Gain\\retention (\%)} \\
\midrule
Pre-RL & 62.00 & 84.99 & --- & --- \\
\rowcolor{rowshade}
Donor & 76.20 & 90.98 & 100.00 & 100.00 \\
rank-1 & 75.80 & 90.74 & 99.48 & 97.18 \\
\rowcolor{rowshade}
rank-2 & 75.40 & 90.60 & 98.95 & 94.37 \\
rank-4 & 76.00 & 90.82 & 99.74 & 98.59 \\
\rowcolor{rowshade}
rank-8 & 75.60 & 90.67 & 99.21 & 95.77 \\
\bottomrule
\end{tabular}
\endgroup
\par\vspace{2pt}
\begin{minipage}{\linewidth}\footnotesize For MATH, rank-1 recovers $(75.80-62.00)/(76.20-62.00)=97.18\%$ of the donor improvement. Retention is reported as a reconstruction-fidelity ratio.\end{minipage}
\end{table}

%% file: tables/T9.tex
\begin{table}[H]
\centering
\caption{\textbf{H1 donor progression.} Stage labels refer to checkpoints within one RL training trajectory.}
\label{tab:T9}
\begingroup
\fontsize{8.0}{10.0}\selectfont
\setlength{\tabcolsep}{4pt}
\renewcommand{\arraystretch}{1.12}
\begin{tabular}{@{}lrrrrrrr@{}}
\toprule
\rowcolor{headerfill}
Stage & \shortstack{Source\\MATH} & \shortstack{MathVista\\Full} & \shortstack{MathVista\\Rank-1} & \shortstack{DynaMath\\Full} & \shortstack{DynaMath\\Rank-1} & \shortstack{MMStar\\Full} & \shortstack{MMStar\\Rank-1} \\
\midrule
Early & 69.60 & 75.80 & 77.90 & 65.47 & 66.91 & 55.33 & 56.47 \\
\rowcolor{rowshade}
Middle & 75.60 & 76.60 & 79.90 & 65.87 & 68.72 & 55.80 & 58.33 \\
Late & 76.20 & 76.70 & 81.00 & 66.07 & 70.06 & 55.93 & 59.67 \\
\bottomrule
\end{tabular}
\endgroup
\end{table}

%% file: tables/T8.tex
\begin{table}[H]
\centering
\caption{\textbf{Transfer from different post-training sources.} Within each source, Full and Rank-1 share the donor and support.}
\label{tab:T8}
\begingroup
\fontsize{8.5}{10.5}\selectfont
\setlength{\tabcolsep}{4pt}
\renewcommand{\arraystretch}{1.12}
\begin{tabular}{@{}lrrrrr@{}}
\toprule
\rowcolor{headerfill}
Source & Transfer & MathVista & DynaMath & MMStar & Median $\rho_1$ \\
\midrule
RL & Full & 76.70 & 66.07 & 55.93 & --- \\
\rowcolor{rowshade}
RL & Rank-1 & 81.00 & 70.06 & 59.67 & 0.18 \\
Offline SFT & Full & 77.10 & 66.67 & 56.27 & --- \\
\rowcolor{rowshade}
Offline SFT & Rank-1 & 78.00 & 67.15 & 56.13 & 0.10 \\
Offline distillation & Full & 77.60 & 67.27 & 56.67 & --- \\
\rowcolor{rowshade}
Offline distillation & Rank-1 & 78.80 & 68.18 & 56.80 & 0.12 \\
On-policy distillation & Full & 77.20 & 67.07 & 56.53 & --- \\
\rowcolor{rowshade}
On-policy distillation & Rank-1 & 79.80 & 69.26 & 58.40 & 0.15 \\
\bottomrule
\end{tabular}
\endgroup
\end{table}

%% file: tables/donor_seed.tex
\begin{table}[H]
\centering
\caption{\textbf{Donor-training seed robustness.} Each seed compares Full and Selective with the same trained donor.}
\label{tab:donor_seed}
\begingroup
\fontsize{8.7}{10.7}\selectfont
\setlength{\tabcolsep}{4pt}
\renewcommand{\arraystretch}{1.12}
\begin{tabular}{@{}lrrrrr@{}}
\toprule
\rowcolor{headerfill}
Seed & \shortstack{Full\\MathVista} & \shortstack{Selective\\MathVista} & \shortstack{MathVista\\gain} & \shortstack{DynaMath\\gain} & \shortstack{MMStar\\gain} \\
\midrule
17 & 76.60 & 80.70 & \gain{4.10} & \gain{3.75} & \gain{3.47} \\
\rowcolor{rowshade}
42 & 76.70 & 81.00 & \gain{4.30} & \gain{3.99} & \gain{3.73} \\
73 & 76.50 & 80.50 & \gain{4.00} & \gain{3.81} & \gain{3.33} \\
\rowcolor{rowshade}
\textbf{Mean $\pm$ SD} & $76.60\pm0.10$ & $80.73\pm0.25$ & $4.13\pm0.15$ & $3.85\pm0.12$ & $3.51\pm0.20$ \\
\bottomrule
\end{tabular}
\endgroup
\par\vspace{2pt}
\begin{minipage}{\linewidth}\footnotesize Gains are in percentage points; standard deviations are sample SDs over the three donor seeds.\end{minipage}
\end{table}

%% file: tables/T13.tex
\begin{table}[H]
\centering
\caption{\textbf{Paired H1 outcomes and 95\% confidence intervals.} Full is the baseline.}
\label{tab:T13}
\begingroup
\fontsize{8.8}{10.8}\selectfont
\setlength{\tabcolsep}{4pt}
\renewcommand{\arraystretch}{1.12}
\begin{tabular}{@{}lrrrrrrr@{}}
\toprule
\rowcolor{headerfill}
Task & $a$ & $b$ & $c$ & $d$ & $N$ & Gain (pp) & 95\% CI \\
\midrule
MathVista & 727 & 40 & 83 & 150 & 1000 & \gain{4.30} & [2.14, 6.46] \\
\rowcolor{rowshade}
MathVerse & 410 & 43 & 110 & 437 & 1000 & \gain{6.70} & [4.31, 9.09] \\
MathVision & 108 & 15 & 41 & 140 & 304 & \gain{8.55} & [3.82, 13.29] \\
\rowcolor{rowshade}
DynaMath & 3197 & 113 & 313 & 1387 & 5010 & \gain{3.99} & [2.10, 5.86] \\
MMStar & 800 & 39 & 95 & 566 & 1500 & \gain{3.73} & [2.23, 5.23] \\
\bottomrule
\end{tabular}
\endgroup
\par\vspace{2pt}
\begin{minipage}{\linewidth}\footnotesize $a$: both correct; $b$: Full only; $c$: Selective only; $d$: both wrong. Gains use integer counts before rounding. DynaMath uses seed-cluster bootstrap; the other tasks use paired normal intervals.\end{minipage}
\end{table}

%% file: tables/dyna_seed.tex
\begin{table}[H]
\centering
\caption{\textbf{DynaMath robustness across variants.} Each of the 501 seed questions has ten variants.}
\label{tab:dyna_seed}
\begingroup
\fontsize{9}{11}\selectfont
\setlength{\tabcolsep}{4pt}
\renewcommand{\arraystretch}{1.12}
\begin{tabular}{@{}lrr@{}}
\toprule
\rowcolor{headerfill}
Statistic & Full & \ourmethod{} \\
\midrule
All-ten-correct (\%) & 29.1 & 34.3 \\
\rowcolor{rowshade}
Median correct variants & 7.0 & 8.0 \\
\bottomrule
\end{tabular}
\endgroup
\par\vspace{2pt}
\begin{minipage}{\linewidth}\footnotesize All-ten-correct is a seed-level metric, distinct from mean accuracy across 5,010 variants. The ten outcomes within a seed are retained individually.\end{minipage}
\end{table}

%% file: tables/T11.tex
\begin{table}[H]
\centering
\caption{\textbf{Generation diagnostics on H1 MathVista.} Token counts and events over 1,000 examples.}
\label{tab:T11}
\begingroup
\fontsize{9}{11}\selectfont
\setlength{\tabcolsep}{4pt}
\renewcommand{\arraystretch}{1.12}
\begin{tabular}{@{}lrrrrr@{}}
\toprule
\rowcolor{headerfill}
Method & \shortstack{Mean\\tokens} & \shortstack{P95\\tokens} & Repeat & \shortstack{Extract\\fail} & \shortstack{At\\cap} \\
\midrule
Receiver & 460 & 1550 & 8 & 5 & 6 \\
\rowcolor{rowshade}
Full & 575 & 2200 & 15 & 8 & 14 \\
IP-Merging & 590 & 2050 & 11 & 6 & 10 \\
\rowcolor{rowshade}
FRISM & 640 & 2200 & 7 & 4 & 8 \\
STAR & 520 & 1820 & 8 & 4 & 7 \\
\rowcolor{rowshade}
AdaRank & 495 & 1740 & 6 & 4 & 6 \\
\rowcolor{oursfill}
SELECTIVE-RL & 430 & 1350 & 3 & 2 & 3 \\
\bottomrule
\end{tabular}
\endgroup
\par\vspace{2pt}
\begin{minipage}{\linewidth}\footnotesize Repetition, extraction failure, and reaching the token cap may overlap.\end{minipage}
\end{table}

%% file: tables/error_types.tex
\begin{table}[H]
\centering
\caption{\textbf{Error categories among 83 Selective-only MathVista successes.}}
\label{tab:error_types}
\begingroup
\fontsize{9}{11}\selectfont
\setlength{\tabcolsep}{4pt}
\renewcommand{\arraystretch}{1.12}
\begin{tabular}{@{}lrr@{}}
\toprule
\rowcolor{headerfill}
Full error category & Count & Share (\%) \\
\midrule
Repetition / loop & 8 & 9.6 \\
\rowcolor{rowshade}
Token-cap truncation & 6 & 7.2 \\
Extraction pathology & 4 & 4.8 \\
\rowcolor{rowshade}
Reasoning / perception & 65 & 78.3 \\
\bottomrule
\end{tabular}
\endgroup
\par\vspace{2pt}
\begin{minipage}{\linewidth}\footnotesize The categories are mutually exclusive; rounded shares may sum to 99.9\%.\end{minipage}
\end{table}

%% file: tables/evaluator.tex
\begin{table}[H]
\centering
\caption{\textbf{Final-answer/evaluator audit on H1 MathVista.} Each model has 200 manually audited examples.}
\label{tab:evaluator}
\begingroup
\fontsize{8.8}{10.8}\selectfont
\setlength{\tabcolsep}{4pt}
\renewcommand{\arraystretch}{1.12}
\begin{tabular}{@{}lrrr@{}}
\toprule
\rowcolor{headerfill}
Statistic & Receiver & Full & \ourmethod{} \\
\midrule
Audited examples & 200.0 & 200.0 & 200.0 \\
\rowcolor{rowshade}
Answer/extraction disagreement & 6.0 & 7.0 & 5.0 \\
False-positive extraction & 2.0 & 3.0 & 2.0 \\
\rowcolor{rowshade}
False-negative extraction & 4.0 & 4.0 & 3.0 \\
Disagreement (\%) & 3.0 & 3.5 & 2.5 \\
\bottomrule
\end{tabular}
\endgroup
\par\vspace{2pt}
\begin{minipage}{\linewidth}\footnotesize The audit is reported separately from the full-benchmark scores.\end{minipage}
\end{table}

%% file: tables/T10.tex
\begin{table}[H]
\centering
\caption{\textbf{Broader H1 capabilities.} TextVQA uses the VQA soft score; other metrics follow Table~\ref{tab:benchmarks}.}
\label{tab:T10}
\begingroup
\fontsize{9}{11}\selectfont
\setlength{\tabcolsep}{4pt}
\renewcommand{\arraystretch}{1.12}
\begin{tabular}{@{}lrrrr@{}}
\toprule
\rowcolor{headerfill}
Benchmark & Receiver & Full & \ourmethod{} & $\Delta$ vs Full \\
\midrule
TextVQA & 82.03 & 81.06 & 83.64 & \gain{2.58} \\
\rowcolor{rowshade}
MMMU & 60.00 & 59.20 & 62.80 & \gain{3.60} \\
OCRBench & 78.60 & 78.00 & 78.70 & \gain{0.70} \\
\rowcolor{rowshade}
POPE & 86.90 & 86.30 & 86.80 & \gain{0.50} \\
\bottomrule
\end{tabular}
\endgroup
\par\vspace{2pt}
\begin{minipage}{\linewidth}\footnotesize OCRBench is reported on a 0--100 scale. POPE changes $86.90\to86.80$ relative to Receiver, so improvements are not uniform across perception metrics.\end{minipage}
\end{table}

%% file: tables/T12.tex
\begin{table}[H]
\centering
\caption{\textbf{Offline construction costs.} Stage times and totals are in seconds; validation search is separate in GPU-hours.}
\label{tab:T12}
\begingroup
\fontsize{8.8}{10.8}\selectfont
\setlength{\tabcolsep}{4pt}
\renewcommand{\arraystretch}{1.12}
\begin{tabular}{@{}lrrrrrr@{}}
\toprule
\rowcolor{headerfill}
Method & SVD & Other & Write & Calib. & Total & \shortstack{Search\\(GPU-h)} \\
\midrule
Receiver & 0 & 0 & 0 & 0 & 0 & 0.0 \\
\rowcolor{rowshade}
Full & 0 & 0 & 35 & 0 & 35 & 3.0 \\
TIES & 0 & 40 & 35 & 0 & 75 & 3.0 \\
\rowcolor{rowshade}
DARE & 0 & 27 & 35 & 0 & 62 & 3.0 \\
TSV & 260 & 90 & 35 & 0 & 385 & 3.0 \\
\rowcolor{rowshade}
STAR & 285 & 75 & 35 & 0 & 395 & 3.0 \\
AdaRank & 300 & 110 & 35 & 1600 & 2045 & 4.0 \\
\rowcolor{rowshade}
IP-Merging & 310 & 175 & 35 & 0 & 520 & 3.0 \\
FRISM & 300 & 195 & 35 & 2400 & 2930 & 4.5 \\
\rowcolor{oursfill}
SELECTIVE-RL & 55 & 0 & 35 & 0 & 90 & 3.0 \\
\bottomrule
\end{tabular}
\endgroup
\end{table}

%% file: tables/T14.tex
\begin{table}[H]
\centering
\caption{\textbf{Deployment footprint.} The merged model retains the recipient architecture.}
\label{tab:T14}
\begingroup
\fontsize{9}{11}\selectfont
\setlength{\tabcolsep}{4pt}
\renewcommand{\arraystretch}{1.12}
\begin{tabular}{@{}lrrr@{}}
\toprule
\rowcolor{headerfill}
Method & Dense ratio & Added module & VRAM (GB) \\
\midrule
Receiver & 1.00 & No & 24.60 \\
\rowcolor{rowshade}
Full & 1.00 & No & 24.60 \\
STAR & 1.00 & No & 24.60 \\
\rowcolor{rowshade}
AdaRank & 1.00 & No & 24.60 \\
FRISM & 1.00 & No & 24.60 \\
\rowcolor{oursfill}
SELECTIVE-RL & 1.00 & No & 24.60 \\
\bottomrule
\end{tabular}
\endgroup
\par\vspace{2pt}
\begin{minipage}{\linewidth}\footnotesize Peak VRAM is reported for batch size 1; completion time is a separate deployment metric.\end{minipage}
\end{table}

%% file: sections/case_appendix.tex
\FloatBarrier
\section{Qualitative Case Audit and Output Analysis}
\label{sec:case-audit}

\paragraph{Selection and duplicate handling.}
The analyzed collection contains 307 records organized for qualitative case analysis: 29 category-1 cases (both comparison models fail), 51 category-2 cases (the receiver succeeds and full merging fails), and 227 category-3 records excluded from the correction examples. The collection is outcome-selected, with other unsuccessful outputs of the proposed method outside its scope. Exact image-and-question matching yields 279 distinct records and 277 distinct decoded images. The displayed panels use distinct problems, with repeated runs grouped by image and question; semantic overlap across datasets may remain. Benchmark-level performance is evaluated separately in the H1--H3 experiments.

The main panels use DynaMath\_4097 (source question 97, variant 5), DynaMath\_7259 (question 259, variant 8), DynaMath\_1314 (question 314, variant 2), MMStar\_920, DynaMath\_4325 (question 325, variant 5), and DynaMath\_8243 (question 243, variant 9). Their decoded image hashes match the retained records. Model predictions, references, question identifiers, and evaluator-extracted answers were retained during the analysis and used to construct the qualitative examples reported in this appendix. Figure text summarizes these records. The retained run records correspond to the Qwen3-VL-8B family receiver and the associated Full and rank-one variants.

\begin{figure}[H]
\centering
\includegraphics[width=\linewidth]{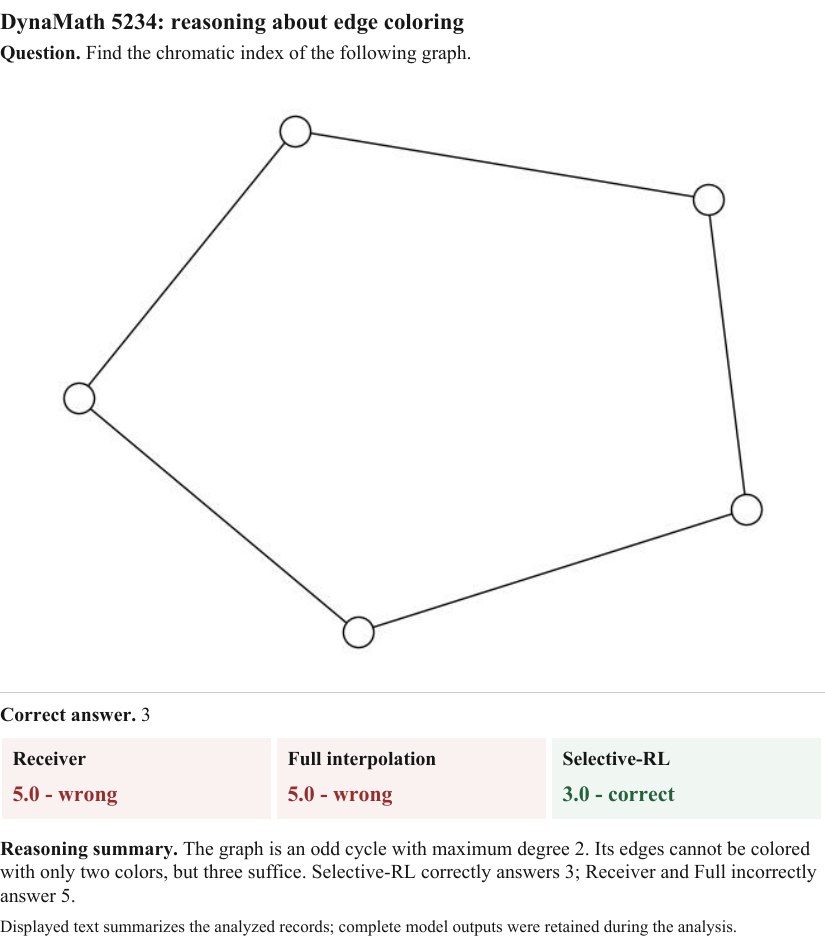}
\caption{\textbf{An edge-coloring correction (DynaMath 5234; question 234, variant 6).} The graph is a five-vertex cycle, whose chromatic index is 3. The receiver and Full each answer 5. Rank-one merging correctly identifies that the odd cycle requires three edge colors.}
\label{fig:case-edge-coloring}
\end{figure}

\clearpage
\paragraph{Preserving correct receiver behavior.}
The following case illustrates a Full-interpolation error while the receiver and rank-one variant both answer correctly. Table~\ref{tab:evaluator} reports the separate H1 audit on 200 examples per model.

\begin{figure}[H]
\centering
\includegraphics[width=\linewidth]{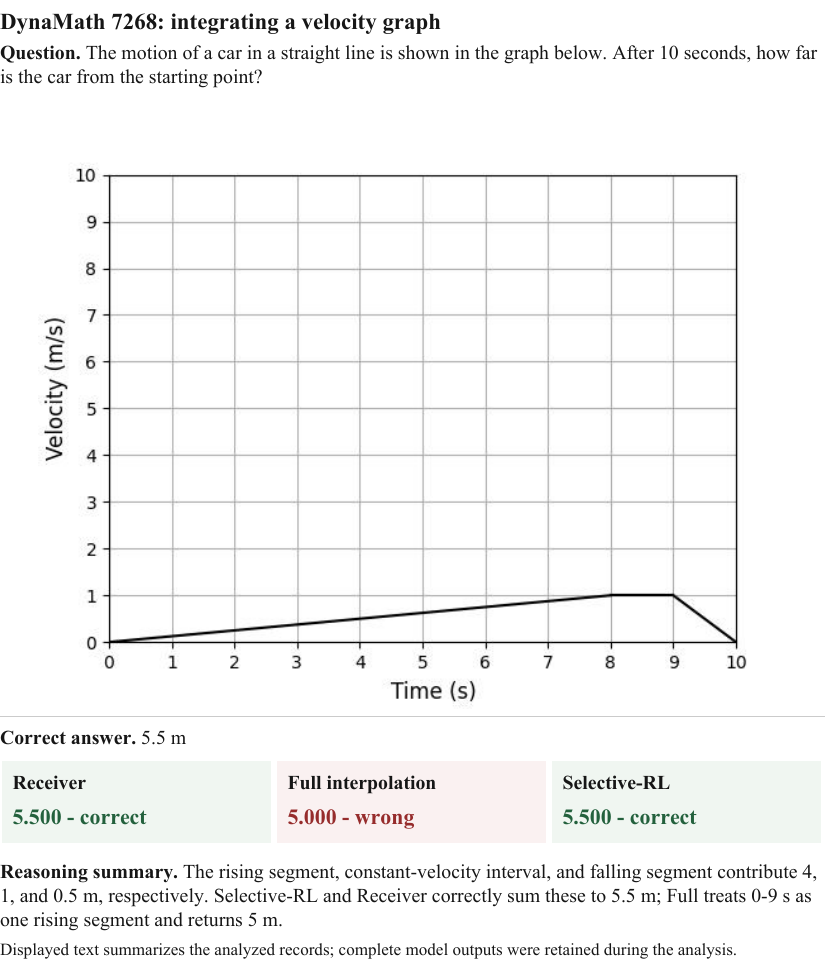}
\caption{\textbf{Correct integration of a velocity graph (DynaMath 7268; question 268, variant 8).} The three segment areas give $4+1+0.5=5.5$ m. Full treats $0$--$9$ s as a single rising segment and answers $5$ m. Rank-one merging and the receiver both account for the constant-velocity interval and correctly answer $5.5$ m.}
\label{fig:case-motion}
\end{figure}